\documentclass[a4paper,fleqn]{cas-dc}

\usepackage[numbers]{natbib}
\usepackage{float}
\usepackage{algorithm}
\usepackage{algpseudocode}
\usepackage{microtype}
\algrenewcommand{\alglinenumber}[1]{\normalfont\fontsize{8}{10}\selectfont #1:}
\newtheorem{theorem}{Theorem}
\newtheorem{lemma}{Lemma}
\newproof{proof}{Proof}

\def\tsc#1{\csdef{#1}{\textsc{\lowercase{#1}}\xspace}}
\tsc{WGM}
\tsc{QE}
\tsc{EP}
\tsc{PMS}
\tsc{BEC}
\tsc{DE}

\makeatletter
\AddToHook{enddocument/afterlastpage}{%
  \immediate\write\@auxout{\string\csxdef{lastpage}{\the\ReadonlyShipoutCounter}}%
}
\makeatother

\begin{document}
\let\WriteBookmarks\relax
\renewcommand{\printorcid}{}
\def\floatpagepagefraction{1}
\def\textpagefraction{.001}
\shorttitle{Blackknife}
\shortauthors{Honglin Gao et~al.}

\title [mode = title]{Blackknife: Query-Free Black-Box Rewiring Attacks on Heterogeneous Graph Neural Networks}                      

\author[1]{Honglin Gao}
\ead{Honglin001@e.ntu.edu.sg}

\author[1]{Junhao Ren}
\ead{JUNHAO002@e.ntu.edu.sg}

\author[1]{Lan Zhao}
\ead{ZHAO0468@e.ntu.edu.sg}

\author[2]{Yue Yang}
\ead{yangyue@cnic.cn}

\author[1]{Peilin Du}
\ead{DUPE0002@e.ntu.edu.sg}

\author[1]{Jindong Chang}
\ead{JINDONG001@e.ntu.edu.sg}

\author[1]{Gaoxi Xiao}
\cormark[1]
\ead{EGXXiao@ntu.edu.sg}

\affiliation[1]{organization={Nanyang Technological University},
                addressline={50 Nanyang Avenue},
                city={Singapore},
                postcode={639798},
                country={Singapore}}

\affiliation[2]{organization={Computer Network Information Center, Chinese Academy of Sciences},
                city={Beijing},
                country={China}}

\cortext[cor1]{Corresponding author}

\begin{abstract}
   Heterogeneous graph neural networks (HGNNs) perform well on graphs with multiple node and relation types, but their robustness to attacks with restricted structural access remains insufficiently understood. We propose Blackknife, a query-free, strict black-box evasion attack that observes only the one-hop structures of the attacked nodes, their ground-truth labels, and raw features within the permitted action space. No victim-model outputs are used to train the surrogate, generate perturbations, or select attacks.Blackknife trains a shared surrogate using only observable edges and relation types, then optimizes structural perturbations through continuous relaxation and projected gradient ascent. Each rewiring operation deletes one edge and adds another of the same relation type, preserving the target node's total degree and degree for each relation type. A cosine-similarity penalty and an adaptive structural classifier further regularize the perturbations through bilevel optimization. Experiments on different datasets demonstrate competitive attack success rates across five victim architectures and three defense settings. The results expose HGNN vulnerabilities under jointly restricted graph access and query-free attack generation.
\end{abstract}


\maketitle

\section{Introduction}
Graph-structured data describe entities and their interactions in social, biological, and knowledge networks \cite{gnnandsocial,gnnandbiological,gnnandknowgraph}. Graph neural networks (GNNs) \cite{gcn,graphsage,gat} use these interactions to learn representations for node classification, link prediction, and graph-level prediction. Many applications also involve multiple entity and relation types. Heterogeneous graphs represent this diversity explicitly and are widely used in financial analysis and bibliographic mining \cite{HGNNandFinance,hgnnFinance3,hgnnCitation3}.

Heterogeneous graph neural networks (HGNNs), including HAN \cite{HAN}, HGT \cite{HGT}, and SimpleHGN \cite{SimpleHGN}, integrate node and relation semantics into message passing. Their predictions nevertheless depend on graph structure, making them vulnerable to adversarial perturbations. By changing a small number of edges, an attacker can alter the messages received by a target node and induce misclassification \cite{nettack,HGAttack,GHAttack}.

Existing heterogeneous graph attacks make different assumptions about information access. Some require access to the full graph structure \cite{HGAttack,GHAttack}, while others rely on querying the victim model's outputs \cite{rl-s2v}. However, such information is often unavailable in practical scenarios. Moreover, independently adding or deleting edges may alter node degrees and introduce noticeable structural changes, increasing the risk of detection. We therefore investigate HGNN vulnerabilities under restricted structural access and without victim model feedback during attack generation.

In our setting, the attacker can observe only the edges directly connected to a set of target nodes and knows their true labels. The attacker can also access the raw features of the targets, their existing neighbors, and a limited set of candidate nodes for new connections. Other connections of these nodes remain unknown. During attack generation, the attacker cannot access the victim model's architecture, parameters, gradients, or outputs.

These constraints introduce three challenges. First, the attacker must identify effective optimization directions without victim feedback. Second, limited local observations make it difficult to capture sufficient structural information for effective attack optimization. Third, structural perturbations must preserve heterogeneous relation semantics while avoiding noticeable changes in local structural statistics that may increase the risk of detection.

\begin{figure}[pos=htbp]
   \centering
   \includegraphics[width=\columnwidth]{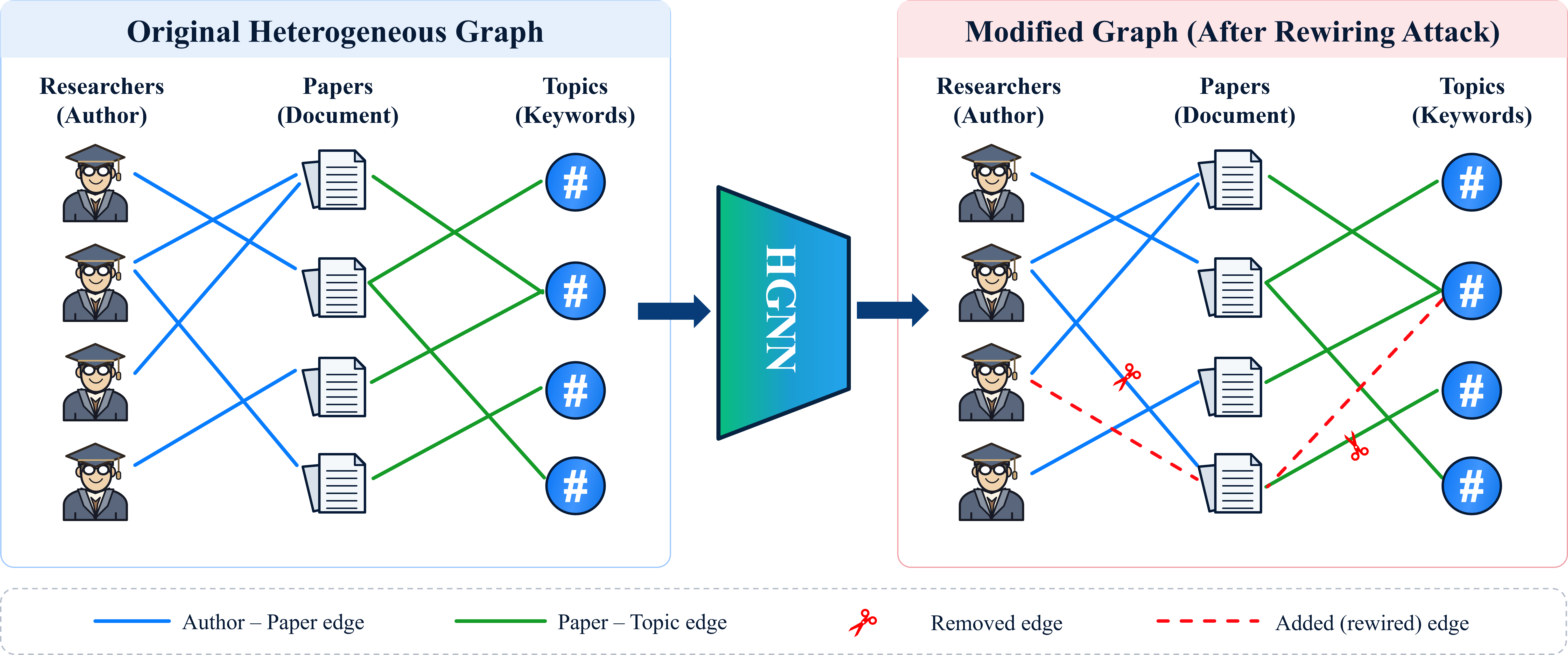}
   \caption{A relation preserving rewiring attack on a heterogeneous bibliographic graph.}
   \label{fig:rewire_attack}
\end{figure}

To address these challenges, we propose Blackknife. A shared local surrogate is trained using the union of observed one hop edges and the ground truth labels of target nodes, while accounting for different relation types. The attacker then optimizes continuous edge selection variables and converts them into pairs of edge deletions and additions within each relation type. This rewiring process preserves both the total degree and the degree associated with each relation type for every target node. Fig.~\ref{fig:rewire_attack} illustrates such a rewiring attack on a bibliographic heterogeneous graph. Furthermore, Blackknife incorporates cosine similarity regularization and an adaptive structural classifier into a bilevel optimization framework to balance attack effectiveness and structural stealthiness.

Our main contributions are as follows:
\begin{itemize}
    \item We formulate a strict black box evasion setting that combines limited one hop structural observations, known target labels and permitted raw features, and no victim model feedback during attack generation.

    \item We propose Blackknife, which trains a shared local surrogate to guide structural rewiring while preserving the total and relation specific degrees of target nodes.

    \item We introduce a bilevel optimization framework with similarity and adaptive structural regularization to encourage detection evasion. Extensive experiments across three datasets and five victim architectures demonstrate the effectiveness of Blackknife against different attack baselines and defense mechanisms.
\end{itemize}

The remainder of this paper is organized as follows. Section~\ref{sec:related} reviews heterogeneous graph neural networks and adversarial attacks on graphs. Section~\ref{sec:prelim} introduces the notation and formalizes the attack problem. Section~\ref{sec:method} presents Blackknife, including the local surrogate, gradient-guided rewiring, and bilevel optimization. Section~\ref{sec:experiments} reports the empirical evaluation. Section~\ref{sec:limitations} discusses limitations and future work, and Section~\ref{sec:conclusion} concludes the paper.

\begin{table*}[t]
   \centering
   \caption{Comparison between Blackknife and representative graph adversarial attack methods.}
   \label{tab:attack_comparison}
   \begin{tabular*}{\textwidth}{@{\extracolsep{\fill}} lcccc}
   \toprule
   Method & Designed for HGs & One Hop Access Only & Query Free Generation & Target Degree Preserving \\
   \midrule
   Nettack   & $\times$ & $\times$ & $\checkmark$ & $\times$ \\
   RL-S2V    & $\times$ & $\times$ & $\times$ & $\times$ \\
   HGAttack  & $\checkmark$ & $\times$ & $\checkmark$ & $\times$ \\
   GHAttack  & $\checkmark$ & $\times$ & $\checkmark$ & $\times$ \\
   \textbf{Blackknife} & $\checkmark$ & $\checkmark$ & $\checkmark$ & $\checkmark$ \\
   \bottomrule
   \end{tabular*}
   \end{table*}

\section{Related Work}
\label{sec:related}
\subsection{Heterogeneous Graph Neural Networks}
Compared with homogeneous graphs, heterogeneous graphs introduce additional modeling challenges because they contain multiple types of nodes and edges, as well as rich semantic dependencies among them. For tasks such as node classification, models are required not only to aggregate neighborhood information, but also to distinguish different relation types and capture cross-type interactions.

To model such complex structures, various Heterogeneous Graph Neural Networks (HGNNs) have been proposed in recent years \cite{RHGNN,HGCN,MAGNN,RpHGNN}. Representative examples include HAN \cite{HAN}, which employs attention mechanisms based on meta paths to aggregate information along semantically meaningful paths; HGT \cite{HGT}, which adopts a Transformer architecture for message passing across different node and edge types; and SimpleHGN \cite{SimpleHGN}, which efficiently incorporates edge type semantics into message propagation. These models have achieved strong performance on heterogeneous graph learning tasks. However, their robustness against adversarial structural perturbations remains insufficiently explored.

\subsection{Adversarial Attacks on Graphs}

Adversarial attacks have revealed the vulnerability of graph neural networks to structural and feature perturbations. Early studies mainly focused on homogeneous graphs. For example, Nettack~\cite{nettack} demonstrates that carefully designed perturbations can mislead node classification, while RL-S2V~\cite{rl-s2v} employs reinforcement learning to generate black box attacks. However, these methods cannot directly accommodate the diverse node types and relation semantics in heterogeneous graphs.

Adversarial attacks on heterogeneous graphs are more challenging due to their complex schema and semantic constraints. Recent studies have explored the robustness of HGNNs. HGAttack~\cite{HGAttack} proposes a transferable gray box attack using surrogate models and gradient based optimization, while GHAttack~\cite{GHAttack} learns a generative attack model to produce structural perturbations. However, these methods do not explicitly preserve node degrees, potentially introducing noticeable structural changes that increase the risk of detection. Moreover, existing methods do not necessarily consider strict restrictions on both model and graph access. In contrast, Blackknife employs degree preserving rewiring under restricted local structural visibility without querying the victim model during attack generation. Table~\ref{tab:attack_comparison} summarizes the key differences between Blackknife and representative graph adversarial attack methods.

\section{Preliminaries and Problem Definition}
\label{sec:prelim}
In this section, we present the necessary preliminaries for adversarial attacks on heterogeneous graphs and formally define the problem in this paper. The meanings of the main notations used throughout this section are summarized in Table~\ref{tab:notations}.

\begin{table}[width=\linewidth,pos=t]
\caption{Summary of main notations.}
\label{tab:notations}
\begin{tabular*}{\tblwidth}{@{\extracolsep{\fill}} l l @{}}
\toprule
\textbf{Notation} & \textbf{Description} \\
\midrule
$G$ & Heterogeneous graph \\
$\mathcal{V},\mathcal{E},X$ & Nodes, edges, and features \\
$\mathcal{T},\mathcal{R}$ & Node types and edge types \\
$t_p$ & Primary node type \\
$\mathcal{T}_{\mathrm{aux}}$ & Auxiliary node types \\
$v_p,\,v_a$ & Target and auxiliary nodes \\
$f$ & Victim classification model \\
$y_p$ & Ground-truth label of $v_p$ \\
$\Delta\mathcal{E}$ & Edge modification set \\
$G\oplus\Delta\mathcal{E}$ & Perturbed graph \\
$c$ & Rewiring budget (deletion and addition pairs) \\
\bottomrule
\end{tabular*}
\end{table}

\subsection{Preliminaries}
\paragraph{Definition 3.1 (Heterogeneous graph).}
A heterogeneous graph is defined as $G=(\mathcal{V}, \mathcal{E}, X)$, where $\mathcal{V}=\{v_1,v_2,\ldots,v_n\}$ denotes the set of nodes, $\mathcal{E}$ denotes the set of typed edges, and $X\in\mathbb{R}^{|\mathcal{V}|\times d}$ is the node feature matrix with feature dimension $d$. Unlike homogeneous graphs, a heterogeneous graph contains multiple types of nodes or edges. Let $\mathcal{T}=\{t_1,t_2,\ldots,t_T\}$ denote the set of node types. Each node $v\in\mathcal{V}$ is associated with a node type through a mapping function $\phi:\mathcal{V}\rightarrow\mathcal{T}$. For each node type $t\in\mathcal{T}$, we denote by $\mathcal{V}_t=\{v\in\mathcal{V}\mid \phi(v)=t\}$ the subset of nodes belonging to type $t$, and by $X_t$ the corresponding feature matrix of nodes in $\mathcal{V}_t$.

The set of edge types is denoted as \(\mathcal{R}=\{r_{t_a,t_b}\mid t_a,t_b\in\mathcal{T}\}\), where each relation type \(r_{t_a,t_b}\) represents edges from nodes of type \(t_a\) to nodes of type \(t_b\). For each relation type \(r_{t_a,t_b}\in\mathcal{R}\), we define a relation-specific adjacency matrix \(A_{r_{t_a,t_b}}\in\{0,1\}^{|\mathcal V|\times|\mathcal V|}\), where \(A_{r_{t_a,t_b}}(i,j)=1\) indicates that there exists an edge from node \(v_i\in\mathcal V_{t_a}\) to node \(v_j\in\mathcal V_{t_b}\) under relation \(r_{t_a,t_b}\), and \(A_{r_{t_a,t_b}}(i,j)=0\) otherwise. Entries corresponding to node-type pairs that do not match the relation schema are set to zero. Accordingly, the edge set can be written as \(\mathcal E=\bigcup_{r_{t_a,t_b}\in\mathcal R}\{(v_i,v_j,r_{t_a,t_b})\mid A_{r_{t_a,t_b}}(i,j)=1\}\). For simplicity, we use \(A_r\) to denote the adjacency matrix associated with relation \(r\).
Therefore, each nonzero entry in \(A_r\) represents a typed edge in \(\mathcal{E}\) under relation \(r\). A graph is considered heterogeneous when it contains more than one node type or more than one edge type, i.e., \(|\mathcal{T}|+|\mathcal{R}|>2\).

\paragraph{Definition 3.2 (Primary and auxiliary node types).}
In a heterogeneous graph, we distinguish between the primary node type and auxiliary node types. The \textit{primary node type}, denoted as $t_p\in\mathcal{T}$, refers to the type of nodes on which classification and attack are performed. The corresponding node set is denoted as $\mathcal{V}_{t_p}$.

Given the primary node type $t_p$, we define the auxiliary node types as all node types directly connected to nodes of type $t_p$. Formally,
$\mathcal{T}_{\mathrm{aux}} = \bigl\{ t_a\in\mathcal{T} \mid \exists\, v_p\in\mathcal{V}_{t_p},\, v_a\in\mathcal{V}_{t_a},\, r\in\mathcal{R},\ (v_p,v_a,r)\in\mathcal{E} \text{ or } (v_a,v_p,r)\in\mathcal{E} \bigr\}$.

\subsection{Problem Definition}

Given a heterogeneous graph $G=(\mathcal{V},\mathcal{E},X)$, a trained node classification model $f$, and a target node $v_p\in\mathcal{V}_{t_p}$ of the primary type $t_p$, we consider an untargeted evasion attack that modifies the graph structure after model training. The goal is to change the prediction of $v_p$ from its true label $y_p$ to any incorrect class.

Structural perturbations are restricted to edges incident to the target and auxiliary-type nodes. Let $\mathcal{E}^{-}_{\mathrm{cand}}(v_p)$ and $\mathcal{E}^{+}_{\mathrm{cand}}(v_p)$ denote the valid deletion and addition candidate sets, respectively. The attack objective is formulated as
\begin{equation}
\begin{aligned}
\max_{\Delta\mathcal{E}}\quad
&\mathbb{1}[f(v_p\mid G\oplus\Delta\mathcal{E})\neq y_p],\\
\mathrm{s.t.}\quad
&\Delta\mathcal{E}^{\pm}\subseteq\mathcal{E}^{\pm}_{\mathrm{cand}}(v_p),\\
&|\Delta\mathcal{E}^{-}_r|=|\Delta\mathcal{E}^{+}_r|,\quad \forall r\in\mathcal{R},\\
&|\Delta\mathcal{E}^{-}|\leq c.
\end{aligned}
\end{equation}

Here, $\Delta\mathcal{E}=\Delta\mathcal{E}^{-}\cup\Delta\mathcal{E}^{+}$ denotes the edge modifications, and $G\oplus\Delta\mathcal{E}$ is the perturbed graph. Each rewiring operation pairs one deletion with one addition of the same relation type, preserving the target node's total and relation-specific degrees. The budget $c$ allows at most $c$ rewiring operations.

\paragraph{Threat model.}
The attacker cannot access the victim's architecture, parameters, gradients, logits, confidence scores, internal representations, or training process. No victim outputs are used during surrogate training, perturbation generation, hyperparameter selection, or attack selection. Only finalized perturbations are submitted to the victim for evaluation.

For each target, the attacker observes its one-hop structure: the target, its directly connected auxiliary neighbors, and the typed edges incident to the target. Other edges of these neighbors are invisible. The attacker knows the true label of each attacked target and can access raw features of the target and nodes in its action space. Addition candidates have known identities, types, and raw features, and are known to be unconnected to the target. Their neighborhoods, degrees, and other structural features are unavailable.

\begin{figure*}[pos=t]
	\centering
	\includegraphics[width=.9\textwidth]{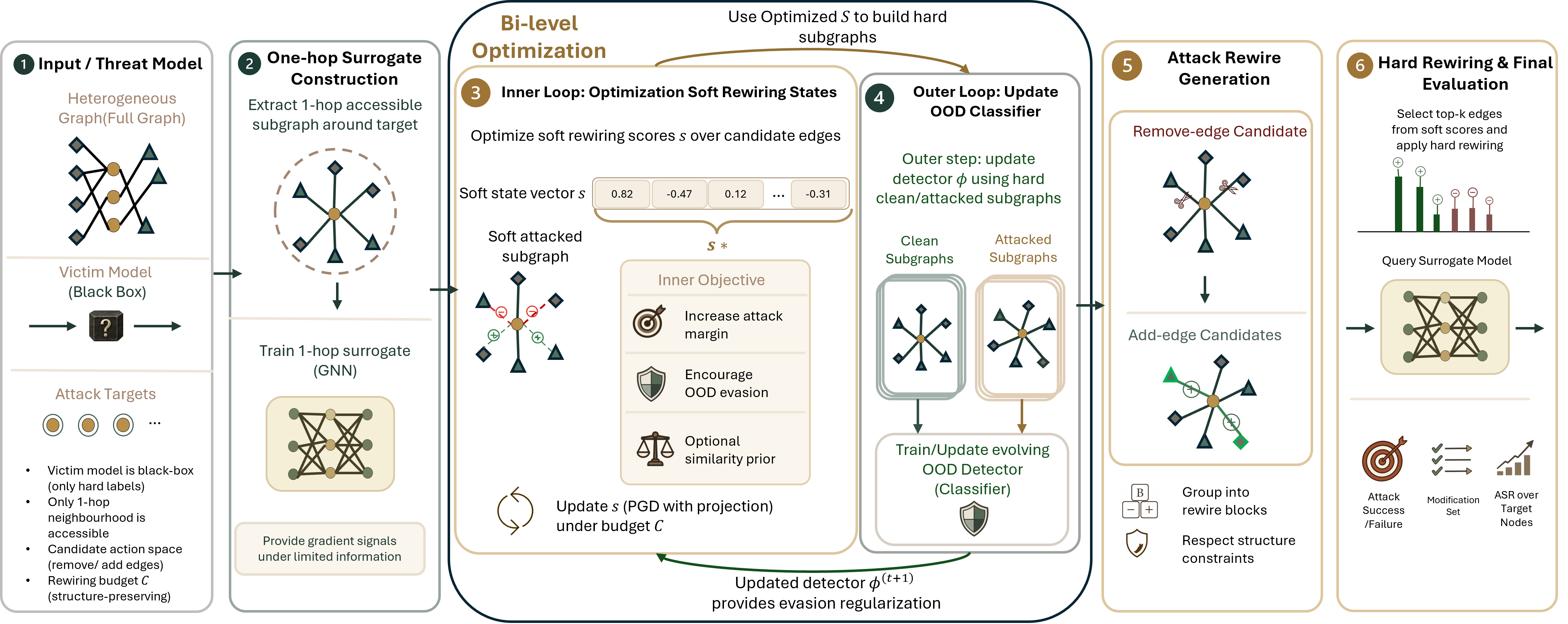}
	\caption{Overall framework of Blackknife for strict black-box attacks on heterogeneous graph neural networks.}
	\label{fig:OverallFramework}
\end{figure*}

\section{Methodology}
\label{sec:method}
In this section, we present the shared local surrogate, the rewiring objective, and the alternating structural-classifier updates. Fig.~\ref{fig:OverallFramework} illustrates local surrogate construction and gradient-guided structural rewiring. Algorithm~\ref{alg:blackknife} generates a separate perturbation set for each target.

\begin{algorithm}[t]
\caption{Blackknife Structural Rewiring}
\label{alg:blackknife}
\begin{algorithmic}[1]
\Require Observed graph $G_{\mathrm{loc}}$, targets $\mathcal{V}_{A}$ with true labels, rewiring budget $c$
\Ensure Separate perturbations $\{\Delta\mathcal{E}_{v_p}\}$
\State Train shared surrogate $g_\theta$ on $G_{\mathrm{loc}}$
\State Initialize internal structural classifier $d_\phi$
\ForAll{$v_p\in\mathcal{V}_{A}$}
\State Construct same-relation deletion and addition candidates
\State Initialize soft rewiring vector $\mathbf{s}_{v_p}$
\EndFor
\For{$t=1,\ldots,T$}
\Statex \textbf{Rewiring update}
\ForAll{$v_p\in\mathcal{V}_{A}$}
\State Recompute deletion similarity references; stop their gradients
\State Compute $\mathcal{J}_{\mathrm{reg}}=\mathcal{J}(v_p)+\lambda_{\mathrm{bl}}\mathcal{R}_{\mathrm{ood}}-\lambda_{\mathrm{cos}}\mathcal{L}_{\mathrm{cos}}$
\State Update $\mathbf{s}_{v_p}$ by projected gradient ascent
\State Enforce bounds, budget, and same-relation pairing
\EndFor
\Statex \textbf{Structural-classifier update}
\State Update $d_\phi$ using clean and discretized attacked local graphs
\EndFor
\ForAll{$v_p\in\mathcal{V}_{A}$}
\State Discretize $\mathbf{s}_{v_p}$ into at most $c$ rewiring pairs
\EndFor
\State \Return $\{\Delta\mathcal{E}_{v_p}\}$
\end{algorithmic}
\end{algorithm}

   \subsection{Local Surrogate Construction}
   
   Under the strict black-box setting, the attacker cannot access the gradients, logits, confidence scores, or internal representations of the victim model. Therefore, directly optimizing structural perturbations on the victim HGNN is infeasible. To address this limitation, Blackknife constructs a local heterogeneous surrogate model using only the structural information observable around the attacked primary nodes, which provides gradient guidance for the subsequent structural rewiring stage.
   
Given attacked primary nodes $\mathcal{V}_{A}\subseteq\mathcal{V}_{t_p}$, let $\mathcal{V}_{\mathrm{acc}}(v_p)$ denote the observed auxiliary neighbors of target $v_p$. Its visible node set is
\begin{equation}
\mathcal{V}_{\mathrm{loc}}(v_p)=\{v_p\}\cup\mathcal{V}_{\mathrm{acc}}(v_p).
\end{equation}
The observed graph for target node \(v_p\) is defined as
\[
G_{v_p}
=
\bigl(
\mathcal{V}_{\mathrm{loc}}(v_p),
\mathcal{E}_{\mathrm{loc}}(v_p),
X_{\mathrm{loc}}(v_p)
\bigr),
\]
where \(\mathcal{E}_{\mathrm{loc}}(v_p)\) is the set of observed typed edges directly connected to \(v_p\). The graph retains node types and raw features.

An attack surrogate is trained on the union of the observed graphs of all target nodes:
\[
G_{\mathrm{loc}}
=
\left(
\bigcup_{v_p\in\mathcal{V}_{A}}\mathcal{V}_{\mathrm{loc}}(v_p),
\;
\bigcup_{v_p\in\mathcal{V}_{A}}\mathcal{E}_{\mathrm{loc}}(v_p),
\;
X_{\mathrm{loc}}
\right).
\]
During rewiring optimization, the surrogate uses candidate nodes’ raw features and represents potential new edges to the target node with continuous weights.

We denote the local surrogate model by $g_{\theta}$. Its message passing mechanism follows the aggregation principle of RGCN \cite{rgcn}, which uses different transformation matrices for different edge types. For a node $v_i$ in $G_{\mathrm{loc}}$, its representation at layer $l+1$ is updated as
\begin{equation}
h_i^{(l+1)}
=
\sigma
\left(
W_0^{(l)}h_i^{(l)}
+
\sum_{r\in\mathcal{R}}
\sum_{v_j\in\mathcal{N}_i^{r}}
\frac{1}{c_{i,r}}
W_r^{(l)}h_j^{(l)}
\right),
\end{equation}
where $h_i^{(0)}$ denotes the initial node features, $\mathcal{N}_i^r$ is the set of neighbors under relation $r$, $W_r^{(l)}$ and $W_0^{(l)}$ are the learnable transformation matrices for relation $r$ and the node itself, respectively, $c_{i,r}$ is a normalization constant, and $\sigma(\cdot)$ is an activation function.
   
After $L$ propagation layers, the surrogate prediction probability for a primary node $v_p$ is given by
\begin{equation}
\hat{\mathbf{p}}_p
=
\mathrm{softmax}
\left(
W_o h_{v_p}^{(L)}
\right),
\end{equation}
where $W_o$ is the output layer parameter and $h_{v_p}^{(L)}$ is the representation of $v_p$. The surrogate is trained by minimizing the cross entropy loss:
\begin{equation}
\theta^{*}
=
\arg\min_{\theta}
\sum_{v_p\in\mathcal{V}_{A}}
\mathcal{L}_{\mathrm{ce}}
\left(
\hat{\mathbf{p}}_p,
y_p
\right).
\end{equation}
   
The local surrogate approximates how structural changes within the observable local subgraph affect the prediction of the target node, thereby providing gradient signals to guide structural rewiring.

An important question is whether the restricted local structure can still support meaningful representation learning. To investigate this, we theoretically analyze how multiple propagation layers aggregate heterogeneous information within a fixed local scope and how the resulting representations depend on local connectivity.
\subsection{Theoretical Analysis of the Local Surrogate}

To investigate this, we follow the idea of decoupling graph scope from message passing depth \cite{shaDowGNN} and consider a normalized linear propagation model over the fixed local heterogeneous subgraph. This formulation allows us to analyze how local connectivity and relation types influence node representations across multiple propagation layers.

Recall that $G_{v_p}$ denotes the observable local subgraph of target node $v_p$, and $\mathcal{R}_{\mathrm{loc}}$ denotes its set of relation types. For each $r\in\mathcal{R}_{\mathrm{loc}}$, let $A_r^{(v_p)}$ denote the adjacency matrix containing only the observed edges of relation $r$ in $G_{v_p}$. We define the relation-weighted adjacency matrix as $B_H=\beta_0 I+\sum_{r\in\mathcal{R}_{\mathrm{loc}}}\beta_r A_r^{(v_p)}$, where $\beta_0>0$ is the self-loop weight and $\beta_r>0$ represents the propagation weight associated with relation $r$. We further define $D_H$ as the diagonal matrix of relation-weighted degrees, where $D_H(u,u)=\sum_w B_H(u,w)$ gives the total propagation weight associated with node $u$. The normalized propagation matrix is then defined as $P_H=D_H^{-1}B_H$.

Let $\widetilde{Z}=H^{(0)}$ collect the initial representations $\{h_u^{(0)}\}$ of all nodes in $G_{v_p}$, and write $\widetilde{z}_u$ for the row corresponding to node $u$. Under this normalized propagation process, the representations after $L$ layers are given by $H^{(L)}=P_H^L\widetilde{Z}$. We use this formulation to characterize how propagation depth and heterogeneous local structure affect the representations obtained within a fixed observation scope.

\begin{lemma}[Fixed Local Scope]
At any depth $L$, $H^{(L)}$ depends only on the fixed local subgraph $G_{v_p}$.
\end{lemma}

Lemma 1 shows that multiple layers of message passing operate entirely within the fixed local subgraph. Regardless of the propagation depth, no additional nodes, edges, or information from outside the observed subgraph are introduced.

\begin{theorem}[Local Convergence]
If $P_H$ is irreducible and aperiodic, it has a unique stationary distribution $\pi_H$ satisfying
\[
\lim_{L\to\infty}[P_H^L]_{v_p,:}=\pi_H^{\mathsf{T}}.
\]
For the target representation $h_{v_p}^{(L)}=[H^{(L)}]_{v_p,:}$, this implies
\[
\lim_{L\to\infty}h_{v_p}^{(L)}
=\pi_H^{\mathsf{T}}\widetilde{Z}
=\sum_{u\in G_{v_p}}\pi_H(u)\widetilde{z}_u.
\]
\end{theorem}

In our setting, the local subgraph supports bidirectional propagation, ensuring that every node is reachable through the target node. Together with the positive self-loop weights, this makes $P_H$ irreducible and aperiodic. Therefore, Theorem 1 establishes that repeated propagation within the fixed local subgraph converges to a stable representation, providing a basis for analyzing how local heterogeneous structure influences the propagation limit.

\begin{theorem}[Local Structural Distinguishability]
\label{thm:local_structure}
Let two local heterogeneous subgraphs share the node set $\mathcal{U}$
and an initial representation matrix $\widetilde{Z}$ with linearly
independent rows. Their relation-weighted normalized propagation
matrices are $P_H$ and $P'_H$. For any integer $L\geq1$, the target
representation is
\[
h_{v_p}^{(L)}
=\sum_{u\in\mathcal{U}}\alpha_{v_p,u}^{(L)}\widetilde{z}_u,
\]
where $\alpha_{v_p,u}^{(L)}=[P_H^L]_{v_p,u}=\sum_{\boldsymbol{x}\in\mathcal{W}_L(v_p,u)}\prod_{\ell=0}^{L-1}P_H(x_\ell,x_{\ell+1})$.
Here, $\mathcal{W}_L(v_p,u)$ contains all length-$L$ directed walks
$\boldsymbol{x}=(x_0,\ldots,x_L)$ from $v_p$ to $u$ within the subgraph.
Repeated nodes and self-loops are allowed, so $L$ may exceed its
observation radius. The other subgraph's representation is defined analogously.
If $[P_H^L]_{v_p,:}\neq[(P'_H)^L]_{v_p,:}$, the two subgraphs produce different target representations after $L$ propagation layers: $h_{v_p}^{(L)}\neq h_{v_p}^{\prime(L)}$.
\end{theorem}

Theorem \ref{thm:local_structure} shows that multilayer propagation combines local node features and relation structure in the target representation. Under the theorem’s conditions, local subgraphs with different aggregation coefficients produce different target representations.
Together, these results show how multilayer propagation combines local
node features with heterogeneous relation structure to form target
representations. They provide a theoretical basis for learning from
fixed local observations and for modifying these representations through
structural rewiring. The proofs are provided in Appendix~\ref{app:proofs} We also evaluate node classification using only one-hop local subgraphs, with the results in Appendix~\ref{app:surrogate} showing that local contexts contain useful class information.

\subsection{Gradient-Guided Structural Rewiring}

After training the local surrogate model, Blackknife uses it to optimize structural rewiring operations. For each attacked primary node $v_p$, we construct candidate sets for edge deletion and addition. Deletion candidates are existing edges between $v_p$ and auxiliary nodes, while addition candidates connect $v_p$ to currently unconnected auxiliary nodes of valid types. Candidates with the same relation type form a group, within which equal numbers of edges are deleted and added to preserve both the heterogeneous schema and the target's relation-specific degrees.

Blackknife transforms discrete edge additions and deletions into a continuous optimization problem through continuous relaxation. For the $i$-th candidate operation, the initial edge state is defined as $b_i \in \{0,1\}$, where $b_i=1$ represents an existing edge (a deletion candidate) and $b_i=0$ denotes a non-existent edge (an addition candidate). By introducing a continuous selection variable $s_i \in [0,1]$, the corresponding soft edge weight is formulated as:
\begin{equation}
    w_i(s_i) = b_i + (1 - 2b_i)s_i.
\end{equation}
This mathematical formulation unifies both graph modifications into a single differentiable framework. Specifically, when $b_i=0$, the weight resolves to $w_i(s_i) = s_i$, meaning that an increase in $s_i$ incrementally adds the candidate edge. Conversely, when $b_i=1$, the weight resolves to $w_i(s_i) = 1 - s_i$, such that increasing $s_i$ incrementally removes the existing edge. By defining operations in this manner, both additions and deletions are uniformly represented and optimized as fully differentiable transformations on a soft graph structure.

On the soft graph, Blackknife computes the logits of $v_p$ using the trained surrogate model and maximizes an untargeted margin objective:
\begin{equation}
\label{eq:margin}
\mathcal{J}(v_p)
=
\max_{k\neq y_p}
z_k(v_p)
-
z_{y_p}(v_p),
\end{equation}
where $z_k(v_p)$ denotes the surrogate logit of node $v_p$ for class $k$. Maximizing this objective reduces the advantage of the true class over the other classes and pushes $v_p$ toward misclassification.

Blackknife updates the continuous selection vector $s$ through projected gradient ascent:
\begin{equation}
s^{(t+1)}
=
\Pi
\left(
s^{(t)}
+
\eta_t
\frac{
\nabla_s \mathcal{J}(v_p)
}{
\left\|
\nabla_s \mathcal{J}(v_p)
\right\|_2
}
\right),
\end{equation}
where $\eta_t$ is the step size and $\Pi(\cdot)$ enforces selection bounds, the rewiring budget, and pairing constraints between edge deletions and additions of the same relation type.

After projected gradient optimization \cite{PGD}, Blackknife discretizes the continuous vector $s$ into a rewiring set $\Delta\mathcal{E}$. It first selects the deletion candidates with the highest scores and then matches each deletion with an addition candidate of the same relation type. The resulting set contains at most $c$ pairs of edge deletions and additions, satisfying $|\Delta\mathcal{E}|\leq 2c$ and equal numbers of deletions and additions for each relation type.

\subsection{Bilevel Optimization Framework}

Although the margin objective in Eq.~\eqref{eq:margin} promotes effective structural perturbations, optimizing attack effectiveness alone may produce conspicuous deviations from the original structure. Blackknife therefore introduces two complementary regularizers: at the microscopic edge level, a similarity loss suppresses additions less similar than the removed edges; at the macroscopic subgraph level, an adaptive classifier encourages the perturbed structure to resemble clean subgraphs.

We introduce similarity regularization to encourage added edges to have feature similarity comparable to that of deleted edges. For each target node $v_p$, deletion and addition candidates of the same relation type are grouped into a block $b\in\mathcal{B}_{v_p}$, with candidate sets $\mathcal{E}_b^-$ and $\mathcal{E}_b^+$, respectively. The similarity of a candidate edge $e=(v_i,v_j,r)$ is computed from the original features of its endpoints as $q_e=\frac{\mathbf{x}_i^{\top}\mathbf{x}_j}{\|\mathbf{x}_i\|_2\|\mathbf{x}_j\|_2}$.

Since the edges to be deleted have not yet been determined during optimization, we compute a reference similarity by averaging the deletion candidates' similarities using softmax weights derived from their current selection scores, $\bar q_b^-=\sum_{e\in\mathcal{E}_b^-}\operatorname{softmax}_{\mathcal{E}_b^-}(s_e)q_e$.
Candidates with higher deletion scores contribute more to this reference. The reference is recomputed at each iteration and treated as a constant when computing the gradient of the similarity loss.

For an addition candidate $e$, the penalty $[\bar q_b^--q_e]_+$ measures how far its similarity falls below the reference, where $[x]_+=\max(x,0)$.
The similarity loss consists of two terms:
\begin{equation}
\begin{aligned}
\mathcal{L}_{\mathrm{cos}}
&=
\frac{1}{|\mathcal{B}_{v_p}|}
\sum_{b\in\mathcal{B}_{v_p}}
\Biggl(
\sum_{e\in\mathcal{E}_{b}^{+}}
\operatorname{softmax}_{\mathcal{E}_{b}^{+}}(s_e)
[\bar q_b^{-}-q_e]_{+}
\\
&\qquad
+
\frac{1}{|\mathcal{E}_{b}^{+}|}
\sum_{e\in\mathcal{E}_{b}^{+}}
s_e[\bar q_b^{-}-q_e]_{+}
\Biggr).
\end{aligned}
\end{equation}
The first term computes the average penalty using the relative selection weights of addition candidates, guiding optimization toward candidates with higher similarity. The second term directly multiplies each penalty by its selection score, discouraging the selection of candidates whose similarity falls below the reference. Together, these terms encourage the addition of edges with higher feature similarity.

To capture richer structural patterns, we introduce a structural classifier $d_{\phi}$ that distinguishes clean local subgraphs from rewired ones. Given the soft perturbed subgraph $\widetilde{G}_{v_p}(\mathbf{s})$, let $\mathcal{R}_{\mathrm{ood}}=\left[d_{\phi}\bigl(\widetilde{G}_{v_p}(\mathbf{s})\bigr)\right]_{0}$, where the subscript $0$ denotes the predicted probability of the clean class. Maximizing $\mathcal{R}_{\mathrm{ood}}$ therefore encourages the generated perturbation to be classified as structurally clean.

Combining the attack objective in Eq.~\eqref{eq:margin} with the two regularization terms, the rewiring variables are optimized according to
\begin{equation}
\max_{\mathbf{s}\in\mathcal{S}}
\;
\mathcal{J}(v_p)
+
\lambda_{\mathrm{bl}}\mathcal{R}_{\mathrm{ood}}
-
\lambda_{\mathrm{cos}}\mathcal{L}_{\mathrm{cos}},
\end{equation}
where $\mathcal{S}$ enforces $s_e\in[0,1]$, the budget of at most $c$ deletion--addition pairs, and same-relation pairing, and $\lambda_{\mathrm{bl}}$ and $\lambda_{\mathrm{cos}}$ control the strengths of the two regularization terms.

The structural classifier is updated using perturbations produced by the current rewiring variables. For each target node $v_p$, Blackknife constructs a clean local subgraph $G_{v_p}$ and a corresponding hard rewired subgraph $G_{v_p}^{\mathrm{adv}}$ obtained by discretizing the current $\mathbf{s}$. The classifier is trained to distinguish the two structures by minimizing
\begin{equation}
\mathcal{L}_{\mathrm{det}}(\phi)
=
-\sum_{v_p\in\mathcal{V}_{A}}
\left[
\log d_{\phi}(G_{v_p})_{0}
+
\log d_{\phi}(G_{v_p}^{\mathrm{adv}})_{1}
\right].
\end{equation}

This interaction leads to the following bilevel formulation:
{\small
\begin{equation}
\begin{aligned}
\max_{\{\mathbf{s}_{v_p}\}}
\quad &
\sum_{v_p\in\mathcal{V}_{A}}
\left[
\mathcal{J}(v_p)
+
\lambda_{\mathrm{bl}}
\mathcal{R}_{\mathrm{ood}}
\left(
\mathbf{s}_{v_p},\phi^{*}
\right)
-
\lambda_{\mathrm{cos}}
\mathcal{L}_{\mathrm{cos}}
\left(
\mathbf{s}_{v_p}
\right)
\right],
\\
\mathrm{s.t.}\quad &
\phi^{*}
=
\arg\min_{\phi}
\mathcal{L}_{\mathrm{det}}
\left(
\phi;
\{\mathbf{s}_{v_p}\}_{v_p\in\mathcal{V}_{A}}
\right),
\\
&
\mathbf{s}_{v_p}\in\mathcal{S}.
\end{aligned}
\end{equation}
}

Inspired by bilevel optimization in graph backdoor attacks \cite{ugba,dpgba}, Blackknife alternately updates the rewiring variables and the structural classifier. In each round, the classifier is held fixed while projected gradient ascent optimizes the soft rewiring vectors. These vectors are then discretized to generate attacked subgraphs, which are used alongside clean subgraphs to update the classifier. The updated classifier provides structural regularization for the next rewiring round, balancing attack effectiveness and structural stealthiness.

\section{Experiments}
\label{sec:experiments}
In this section, we evaluate Blackknife by answering the following five research questions:
\begin{itemize}
    \item \textbf{RQ1:} How effective is Blackknife across different heterogeneous graph datasets and victim models?
    \item \textbf{RQ2:} How well does Blackknife perform against different topology-based defense strategies?
    \item \textbf{RQ3:} How do regularization and learned rewiring contribute to detection evasion and attack effectiveness?
    \item \textbf{RQ4:} How do key hyperparameters affect the attack performance of Blackknife?
    \item \textbf{RQ5:} How do different attacks change the degree distribution of the attacked target nodes?
\end{itemize}

\subsection{Experiment Settings}
\begin{table}[ht]
   \centering
   \small
   \normalfont
   \caption{\normalfont{Dataset Statistics.}}
   \label{tab:dataset statistics}
   \resizebox{\linewidth}{!}{%
   \begin{tabular}{lccccc}
   \toprule
   \textbf{Dataset} & \textbf{\#Node Types} & \textbf{\#Edge Types} & \textbf{\#Nodes} & \textbf{\#Edges} & \textbf{Primary Type} \\
   \midrule
   ACM   & 3 & 4 & 11252  & 34864  & paper  \\
   IMDB  & 3 & 4 & 11616  & 34212  & movie \\
   DBLP  & 4 & 6 & 26198  & 242142 & author  \\
   \bottomrule
   \end{tabular}%
   }
\end{table}
We use three benchmark heterogeneous graphs: ACM, IMDB, and DBLP. ACM classifies papers by research area; IMDB classifies movies by genre, with actors and directors as auxiliary entities; DBLP classifies authors by research area. Table~\ref{tab:dataset statistics} summarizes the datasets.

For each dataset, we randomly select \(10\%\) of nodes of the primary type as attack targets. The shared attack surrogate is trained using all these targets’ true labels and the union of their observed edges within one hop. For each target and auxiliary type, addition candidates are ranked by their mean cosine distance to the target’s existing neighbors of that type. The closest candidates, numbering \(5\%\) of that type’s total population, form the permitted addition pool. Only the target’s connectivity within one hop and the raw features within its action space are visible.

We evaluate five victim architectures: HAN~\cite{HAN}, HGT~\cite{HGT}, SimpleHGN~\cite{SimpleHGN}, RE-GNN~\cite{REGNN}, and SeHGNN~\cite{SeHGNN}. Baselines include HGAttack~\cite{HGAttack}, a transferable heterogeneous graph attack using a surrogate model; GHAttack~\cite{GHAttack}, a generative heterogeneous graph attack; and RL-S2V~\cite{rl-s2v}, a structural attack using reinforcement learning. At the reported budget \(c\), Blackknife uses at most \(c\) pairs of edge deletions and additions, while each baseline is allowed \(2c\) individual edge modifications. All methods are compared under the same allowance for edge modifications.

Targets are evaluated separately. Each target’s finalized changes are applied to the original graph, evaluated, and then reversed before evaluating another target. ASR is the fraction of initially correctly classified targets that become misclassified after attack. Targets already misclassified on the clean graph are excluded from its denominator. Victim predictions on clean and attacked graphs are collected to compute ASR.

We evaluate Blackknife against defenses operating at three levels: edges \cite{GNNGuard,ugba}, local subgraphs \cite{defense_graph_classifier}, and the prediction model \cite{RoHe}. At the edge level, Prune filters added edges based on feature similarity, with a separate cutoff for each relation type. Each cutoff is set to the fifth percentile of similarities along existing edges of that relation in the clean graph. At the subgraph level, an independently trained structural classifier detects attacked local subgraphs. At the model level, RoHe~\cite{RoHe} provides a robust HGNN architecture. Prune and subgraph detection are evaluated using HGT as the victim model.

All experiments are conducted on a workstation equipped with an AMD Ryzen Threadripper PRO 5975WX CPU with 32 physical cores and 64 threads, and an NVIDIA RTX A5000 GPU with 24 GB VRAM. Each experiment is repeated three times with different random seeds, and we report the mean and standard deviation.

\begin{table*}[t]
   \centering
   \caption{Attack success rates under rewiring budgets $c=1,2,3$. Each baseline is allowed $2c$ individual edge modifications.}
   \label{tab:attack_success_rate}
   \resizebox{0.92\textwidth}{!}{
   \small
   \begin{tabular}{lllccccc}
   \toprule
   \textbf{Dataset} & \textbf{Attack Method} & \textbf{Budget $c$} & \textbf{HAN} & \textbf{HGT} & \textbf{SimpleHGN} & \textbf{REGNN} & \textbf{SeHGNN} \\
   \midrule
   \multirow{12}{*}{ACM}
   & HGAttack & \multirow{4}{*}{$1$}
   & $0.1917 \pm 0.0965$
   & $0.1667 \pm 0.0548$
   & $0.1767 \pm 0.0153$
   & $0.1300 \pm 0.0229$
   & $\mathbf{0.7700 \pm 0.1353}$ \\
   & RL-S2V &
   & $0.0700 \pm 0.0150$
   & $0.0733 \pm 0.0275$
   & $0.0433 \pm 0.0029$
   & $0.0367 \pm 0.0115$
   & $0.1567 \pm 0.1020$ \\
   & GHAttack &
   & $0.1800 \pm 0.1262$
   & $0.1733 \pm 0.0535$
   & $0.0883 \pm 0.1098$
   & $0.0600 \pm 0.0300$
   & $0.5250 \pm 0.4277$ \\
   & Blackknife &
   & $\mathbf{0.3517 \pm 0.0539}$
   & $\mathbf{0.3300 \pm 0.0427}$
   & $\mathbf{0.3150 \pm 0.0200}$
   & $\mathbf{0.2350 \pm 0.0218}$
   & $0.3933 \pm 0.0340$ \\
   \cmidrule{2-8}
   & HGAttack & \multirow{4}{*}{$2$}
   & $0.3433 \pm 0.1296$
   & $0.3317 \pm 0.0425$
   & $0.3367 \pm 0.0225$
   & $0.2183 \pm 0.0465$
   & $\mathbf{0.9400 \pm 0.0433}$ \\
   & RL-S2V &
   & $0.1000 \pm 0.0100$
   & $0.1800 \pm 0.1905$
   & $0.0683 \pm 0.0144$
   & $0.0667 \pm 0.0225$
   & $0.5017 \pm 0.0777$ \\
   & GHAttack &
   & $0.1467 \pm 0.0993$
   & $0.3183 \pm 0.1594$
   & $0.0767 \pm 0.0597$
   & $0.0717 \pm 0.0605$
   & $0.5383 \pm 0.3926$ \\
   & Blackknife &
   & $\mathbf{0.5700 \pm 0.0377}$
   & $\mathbf{0.6850 \pm 0.0265}$
   & $\mathbf{0.6567 \pm 0.0161}$
   & $\mathbf{0.5400 \pm 0.0132}$
   & $0.5683 \pm 0.0257$ \\
   \cmidrule{2-8}
   & HGAttack & \multirow{4}{*}{$3$}
   & $0.4283 \pm 0.1457$
   & $0.4083 \pm 0.0144$
   & $0.4317 \pm 0.0569$
   & $0.3000 \pm 0.0350$
   & $\mathbf{0.9517 \pm 0.0231}$ \\
   & RL-S2V &
   & $0.1150 \pm 0.0477$
   & $0.0983 \pm 0.0153$
   & $0.1250 \pm 0.0954$
   & $0.0767 \pm 0.0202$
   & $0.1283 \pm 0.0202$ \\
   & GHAttack &
   & $0.2967 \pm 0.2255$
   & $0.2350 \pm 0.0740$
   & $0.2117 \pm 0.2394$
   & $0.1450 \pm 0.1394$
   & $0.7900 \pm 0.1370$ \\
   & Blackknife &
   & $\mathbf{0.7633 \pm 0.0029}$
   & $\mathbf{0.8433 \pm 0.0225}$
   & $\mathbf{0.8250 \pm 0.0218}$
   & $\mathbf{0.7417 \pm 0.0153}$
   & $0.7550 \pm 0.0265$ \\
   \midrule
   \multirow{12}{*}{DBLP}
   & HGAttack & \multirow{4}{*}{$1$}
   & $0.3050 \pm 0.0100$
   & $0.2383 \pm 0.0076$
   & $0.2500 \pm 0.0087$
   & $0.2017 \pm 0.0189$
   & $0.1550 \pm 0.0218$ \\
   & RL-S2V &
   & $0.2917 \pm 0.0208$
   & $0.1433 \pm 0.0076$
   & $0.1467 \pm 0.0284$
   & $0.1867 \pm 0.0301$
   & $0.2800 \pm 0.0229$ \\
   & GHAttack &
   & $0.2333 \pm 0.0115$
   & $0.2933 \pm 0.0058$
   & $0.1983 \pm 0.0586$
   & $\mathbf{0.4517 \pm 0.0058}$
   & $0.1800 \pm 0.0000$ \\
   & Blackknife &
   & $\mathbf{0.4250 \pm 0.0050}$
   & $\mathbf{0.4283 \pm 0.0126}$
   & $\mathbf{0.4033 \pm 0.0104}$
   & $0.3817 \pm 0.0351$
   & $\mathbf{0.5783 \pm 0.0351}$ \\
   \cmidrule{2-8}
   & HGAttack & \multirow{4}{*}{$2$}
   & $0.3550 \pm 0.0087$
   & $0.2983 \pm 0.0176$
   & $0.3033 \pm 0.0076$
   & $0.2483 \pm 0.0225$
   & $0.2017 \pm 0.0161$ \\
   & RL-S2V &
   & $0.1117 \pm 0.1068$
   & $0.3267 \pm 0.0293$
   & $0.3517 \pm 0.0506$
   & $0.4517 \pm 0.0104$
   & $0.3900 \pm 0.1026$ \\
   & GHAttack &
   & $0.2333 \pm 0.0029$
   & $0.5367 \pm 0.0058$
   & $0.4017 \pm 0.0076$
   & $0.5733 \pm 0.0029$
   & $0.3183 \pm 0.0153$ \\
   & Blackknife &
   & $\mathbf{0.6367 \pm 0.0202}$
   & $\mathbf{0.6050 \pm 0.0278}$
   & $\mathbf{0.5917 \pm 0.0104}$
   & $\mathbf{0.5800 \pm 0.0180}$
   & $\mathbf{0.6983 \pm 0.0153}$ \\
   \cmidrule{2-8}
   & HGAttack & \multirow{4}{*}{$3$}
   & $0.3883 \pm 0.0126$
   & $0.3200 \pm 0.0260$
   & $0.3183 \pm 0.0153$
   & $0.2767 \pm 0.0321$
   & $0.2267 \pm 0.0208$ \\
   & RL-S2V &
   & $0.2283 \pm 0.0225$
   & $0.4767 \pm 0.0535$
   & $0.4567 \pm 0.0379$
   & $0.5183 \pm 0.0321$
   & $0.5000 \pm 0.0687$ \\
   & GHAttack &
   & $0.2517 \pm 0.0126$
   & $0.5917 \pm 0.0076$
   & $0.5300 \pm 0.0087$
   & $0.6133 \pm 0.0076$
   & $0.4250 \pm 0.0132$ \\
   & Blackknife &
   & $\mathbf{0.7250 \pm 0.0482}$
   & $\mathbf{0.6900 \pm 0.0173}$
   & $\mathbf{0.6750 \pm 0.0180}$
   & $\mathbf{0.6650 \pm 0.0087}$
   & $\mathbf{0.7283 \pm 0.0340}$ \\
   \midrule
   \multirow{12}{*}{IMDB}
   & HGAttack & \multirow{4}{*}{$1$}
   & $0.2317 \pm 0.0202$
   & $0.1617 \pm 0.0076$
   & $0.2833 \pm 0.0161$
   & $0.2117 \pm 0.0104$
   & $0.1983 \pm 0.0058$ \\
   & RL-S2V &
   & $0.2633 \pm 0.0701$
   & $0.1533 \pm 0.0275$
   & $0.2133 \pm 0.0520$
   & $0.2117 \pm 0.0511$
   & $0.2367 \pm 0.0362$ \\
   & GHAttack &
   & $0.1633 \pm 0.1575$
   & $0.2083 \pm 0.0577$
   & $0.1517 \pm 0.0683$
   & $0.0500 \pm 0.0000$
   & $0.2267 \pm 0.0388$ \\
   & Blackknife &
   & $\mathbf{0.4983 \pm 0.0161}$
   & $\mathbf{0.4483 \pm 0.0161}$
   & $\mathbf{0.4683 \pm 0.0247}$
   & $\mathbf{0.4167 \pm 0.0126}$
   & $\mathbf{0.4683 \pm 0.0355}$ \\
   \cmidrule{2-8}
   & HGAttack & \multirow{4}{*}{$2$}
   & $0.2850 \pm 0.0200$
   & $0.2150 \pm 0.0132$
   & $0.3783 \pm 0.0115$
   & $0.2850 \pm 0.0100$
   & $0.2667 \pm 0.0176$ \\
   & RL-S2V &
   & $0.2700 \pm 0.1165$
   & $0.1867 \pm 0.0535$
   & $0.2750 \pm 0.0492$
   & $0.2500 \pm 0.0786$
   & $0.2967 \pm 0.0605$ \\
   & GHAttack &
   & $0.4883 \pm 0.1662$
   & $0.4067 \pm 0.0861$
   & $0.2767 \pm 0.0723$
   & $0.2150 \pm 0.0265$
   & $0.4867 \pm 0.2570$ \\
   & Blackknife &
   & $\mathbf{0.5783 \pm 0.0189}$
   & $\mathbf{0.5150 \pm 0.0229}$
   & $\mathbf{0.5417 \pm 0.0161}$
   & $\mathbf{0.4817 \pm 0.0126}$
   & $\mathbf{0.6083 \pm 0.0176}$ \\
   \cmidrule{2-8}
   & HGAttack & \multirow{4}{*}{$3$}
   & $0.3083 \pm 0.0202$
   & $0.2383 \pm 0.0176$
   & $0.4200 \pm 0.0173$
   & $0.3317 \pm 0.0176$
   & $0.3000 \pm 0.0218$ \\
   & RL-S2V &
   & $0.1000 \pm 0.0218$
   & $0.1933 \pm 0.0076$
   & $0.3167 \pm 0.0448$
   & $0.2833 \pm 0.1436$
   & $0.3250 \pm 0.0661$ \\
   & GHAttack &
   & $0.2750 \pm 0.1612$
   & $0.5017 \pm 0.1098$
   & $0.3417 \pm 0.0723$
   & $0.3067 \pm 0.2062$
   & $0.4033 \pm 0.1627$ \\
   & Blackknife &
   & $\mathbf{0.6017 \pm 0.0176}$
   & $\mathbf{0.6300 \pm 0.0173}$
   & $\mathbf{0.5983 \pm 0.0231}$
   & $\mathbf{0.5383 \pm 0.0247}$
   & $\mathbf{0.6950 \pm 0.0260}$ \\
   \bottomrule
   \end{tabular}
   }
\end{table*}

\begin{table}[t]
   \centering
   \caption{Defended ASR at $c=3$. Prune and OOD detection use HGT; RoHe is the robust HGNN setting.}
   \label{tab:defense_results}
   \resizebox{\linewidth}{!}{
   \begin{tabular}{llccc}
   \toprule
   \textbf{Dataset} & \textbf{Attack Method}
   & \textbf{Prune}
   & \textbf{OOD Detector}
   & \textbf{RoHe} \\
   \midrule
   \multirow{4}{*}{ACM}
   & HGAttack
   & $0.3950 \pm 0.0050$
   & $0.2920 \pm 0.0290$
   & $0.3417 \pm 0.0737$ \\
   & RL-S2V
   & $0.1000 \pm 0.0180$
   & $0.1063 \pm 0.0203$
   & $0.1467 \pm 0.0425$ \\
   & GHAttack
   & $0.2333 \pm 0.0769$
   & $0.0680 \pm 0.0830$
   & $0.5600 \pm 0.3018$ \\
   & Blackknife
   & $\mathbf{0.8333 \pm 0.0176}$
   & $\mathbf{0.6470 \pm 0.0280}$
   & $\mathbf{0.6850 \pm 0.0132}$ \\
   \midrule
   \multirow{4}{*}{DBLP}
   & HGAttack
   & $0.2667 \pm 0.0231$
   & $0.2470 \pm 0.0100$
   & $0.2150 \pm 0.0100$ \\
   & RL-S2V
   & $0.4500 \pm 0.0656$
   & $0.3720 \pm 0.1244$
   & $0.4333 \pm 0.0629$ \\
   & GHAttack
   & $0.5917 \pm 0.0076$
   & $0.0620 \pm 0.0030$
   & $0.3133 \pm 0.0321$ \\
   & Blackknife
   & $\mathbf{0.6817 \pm 0.0231}$
   & $\mathbf{0.5100 \pm 0.0350}$
   & $\mathbf{0.6133 \pm 0.0275}$ \\
   \midrule
   \multirow{4}{*}{IMDB}
   & HGAttack
   & $0.0450 \pm 0.0173$
   & $0.0450 \pm 0.0280$
   & $0.2517 \pm 0.0104$ \\
   & RL-S2V
   & $0.0533 \pm 0.0252$
   & $0.2365 \pm 0.0224$
   & $0.2033 \pm 0.0777$ \\
   & GHAttack
   & $0.4533 \pm 0.1042$
   & $0.0930 \pm 0.1620$
   & $0.4217 \pm 0.1710$ \\
   & Blackknife
   & $\mathbf{0.5417 \pm 0.0375}$
   & $\mathbf{0.4480 \pm 0.0120}$
   & $\mathbf{0.6617 \pm 0.0225}$ \\
   \bottomrule
   \end{tabular}
   }
\end{table}

\subsection{Attack Performance (RQ1)}
Table~\ref{tab:attack_success_rate} compares ASR at $c=1,2,3$, with matched allowances of $2c$ individual edge modifications. Blackknife achieves the highest mean ASR in 41 of the 45 evaluated settings. At $c=3$, it outperforms all baselines on every victim model on DBLP and IMDB, and on four of the five models on ACM.

Blackknife’s mean ASR increases with budget across all dataset and victim combinations. These results demonstrate its effectiveness against diverse victim architectures under restricted structural observations and without victim feedback during attack generation, while preserving each target’s total and relation-specific degrees.

\subsection{Defense Resistance (RQ2)}
Table~\ref{tab:defense_results} reports defended ASR at $c=3$. Blackknife has the highest mean in all nine dataset--defense combinations. On ACM, its values are $0.8333$ under Prune, $0.6470$ under OOD detection, and $0.6850$ against RoHe. The corresponding DBLP values are $0.6817$, $0.5100$, and $0.6133$; the IMDB values are $0.5417$, $0.4480$, and $0.6617$.

These results show that the generated perturbations remain effective under the evaluated filtering, detection, and robust-model settings. The independent OOD detector reduces attack effectiveness relative to the undefended HGT results on all three datasets, but Blackknife retains an advantage over the baseline means. Pairing edge deletions with additions preserves node degrees, while the two regularizers address other detectable properties. The ablation results below examine the regularizers’ contributions to evasion.

\subsection{Ablation Study (RQ3)}
\begin{figure*}[t]
   \centering
   \includegraphics[width=\textwidth]{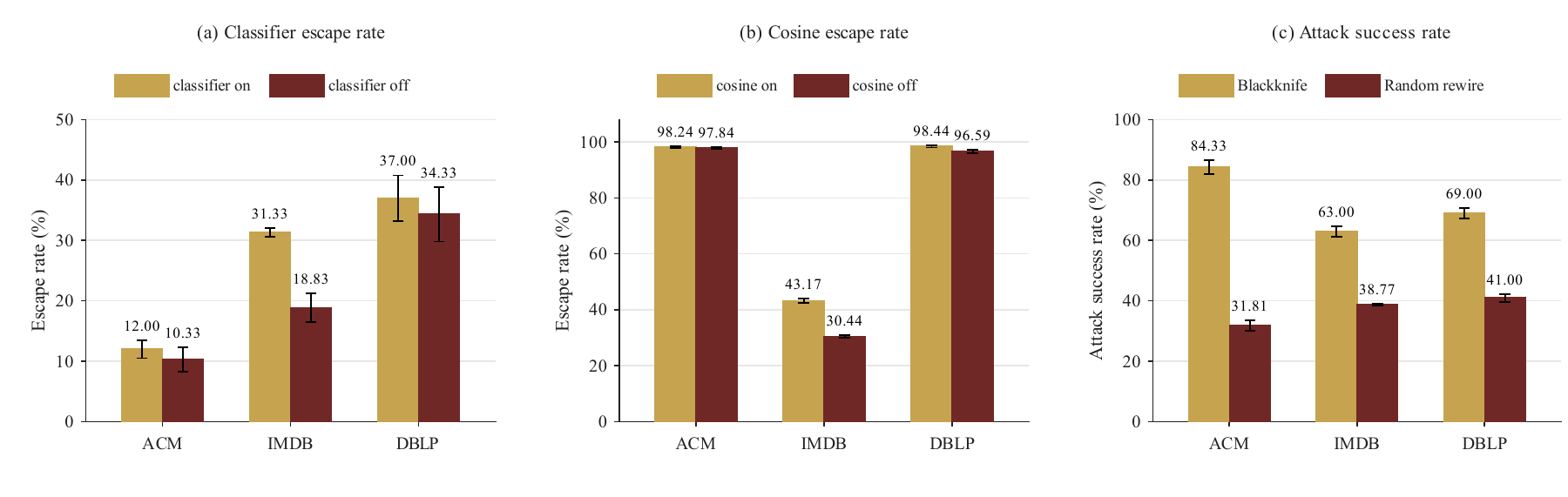}
   \caption{Ablation on HGT at $c=3$: classifier escape with and without the structural classifier (a), cosine escape with and without cosine regularization (b), and attack success rate of Blackknife versus random rewiring (c). The evaluation classifier is independent of the attack classifier.}
   \label{fig:blackknife_ablation}
\end{figure*}

Fig.~\ref{fig:blackknife_ablation} presents three ablations on HGT at $c=3$. For panel (a), we assume that the defender knows the attack modifies the graph through rewiring and uses a separate classifier specifically designed to detect rewiring attacks. Classifier escape rate is the fraction of attacked examples that this detector labels as clean. Panel (a) compares this rate with and without structural classifier regularization during attack generation (classifier on/off). Panel (b) compares cosine escape rates with and without cosine regularization (cosine on/off), while panel (c) compares Blackknife with random rewiring under the same budget.

Structural classifier regularization improves evasion of the rewiring detector on all three datasets, particularly on IMDB. Cosine regularization also provides its largest improvement on IMDB, while escape rates on ACM and DBLP remain high with or without it. Compared with uniformly random rewiring, Blackknife increases ASR by $52.52$, $24.23$, and $28.00$ percentage points on ACM, IMDB, and DBLP, respectively, demonstrating the benefit of optimizing the rewiring operations.

\subsection{Hyperparameter Analysis (RQ4)}
\begin{figure*}[t]
   \centering
   \includegraphics[width=0.24\textwidth]{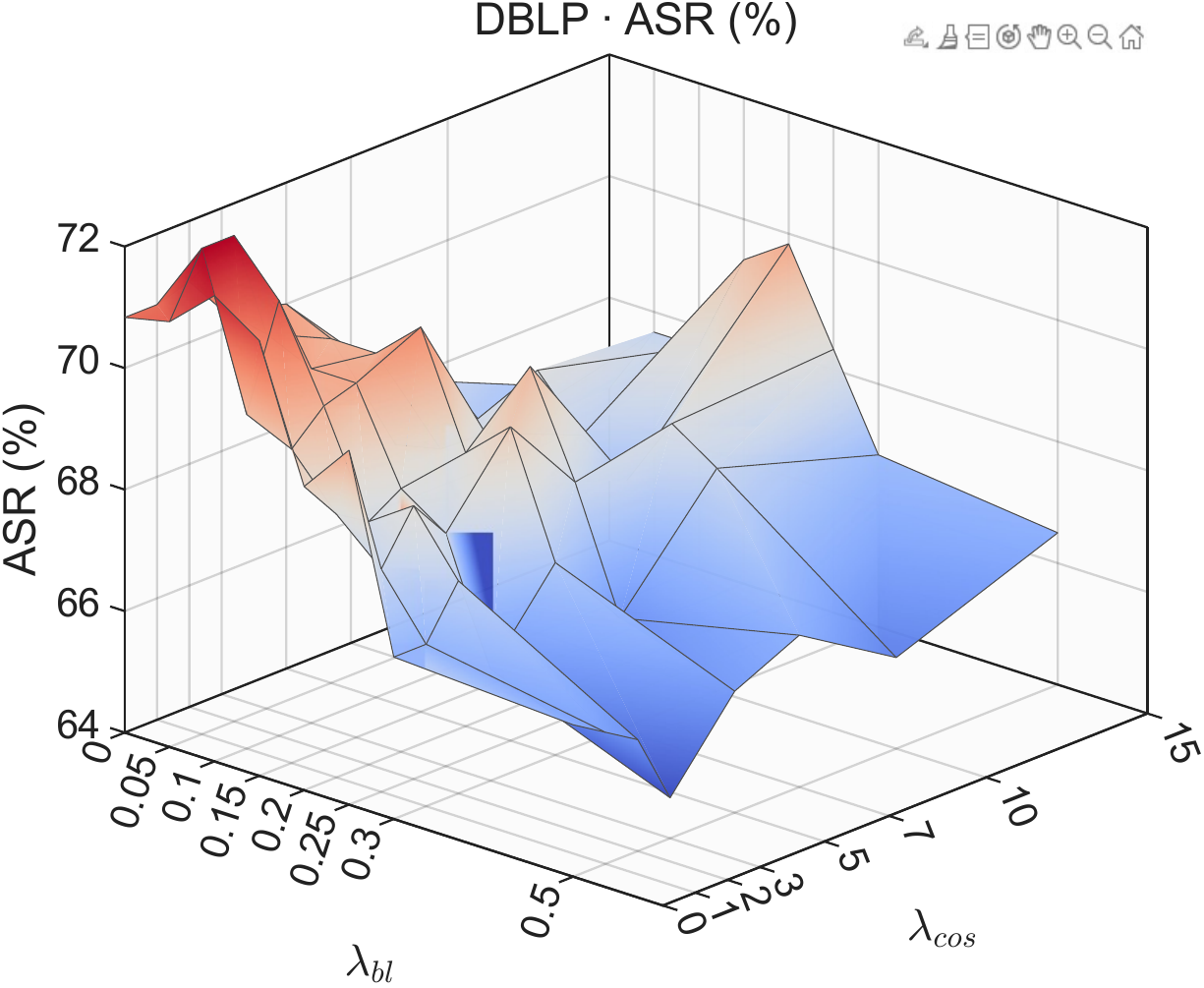}
   \hfill
   \includegraphics[width=0.24\textwidth]{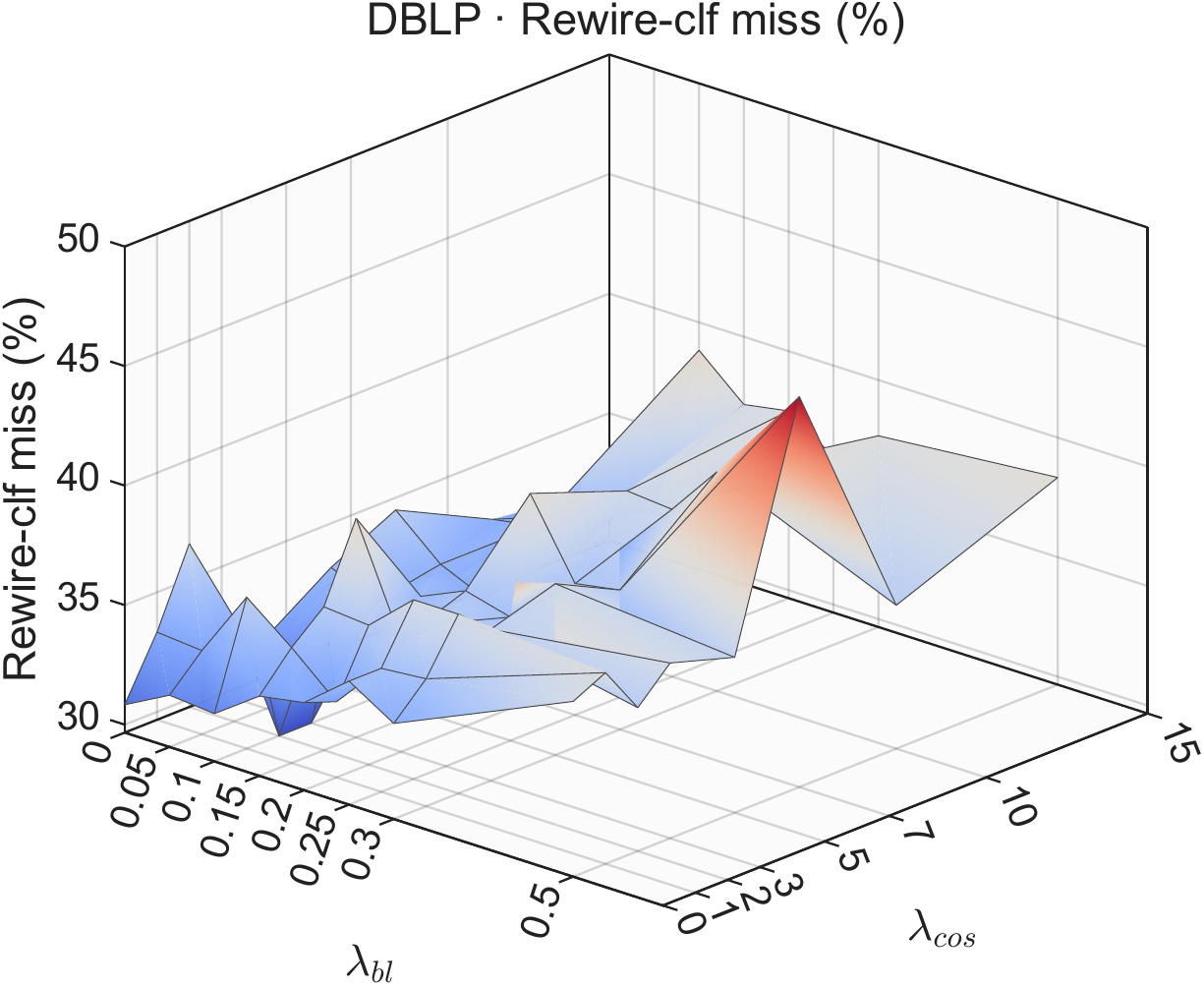}
   \hfill
   \includegraphics[width=0.24\textwidth]{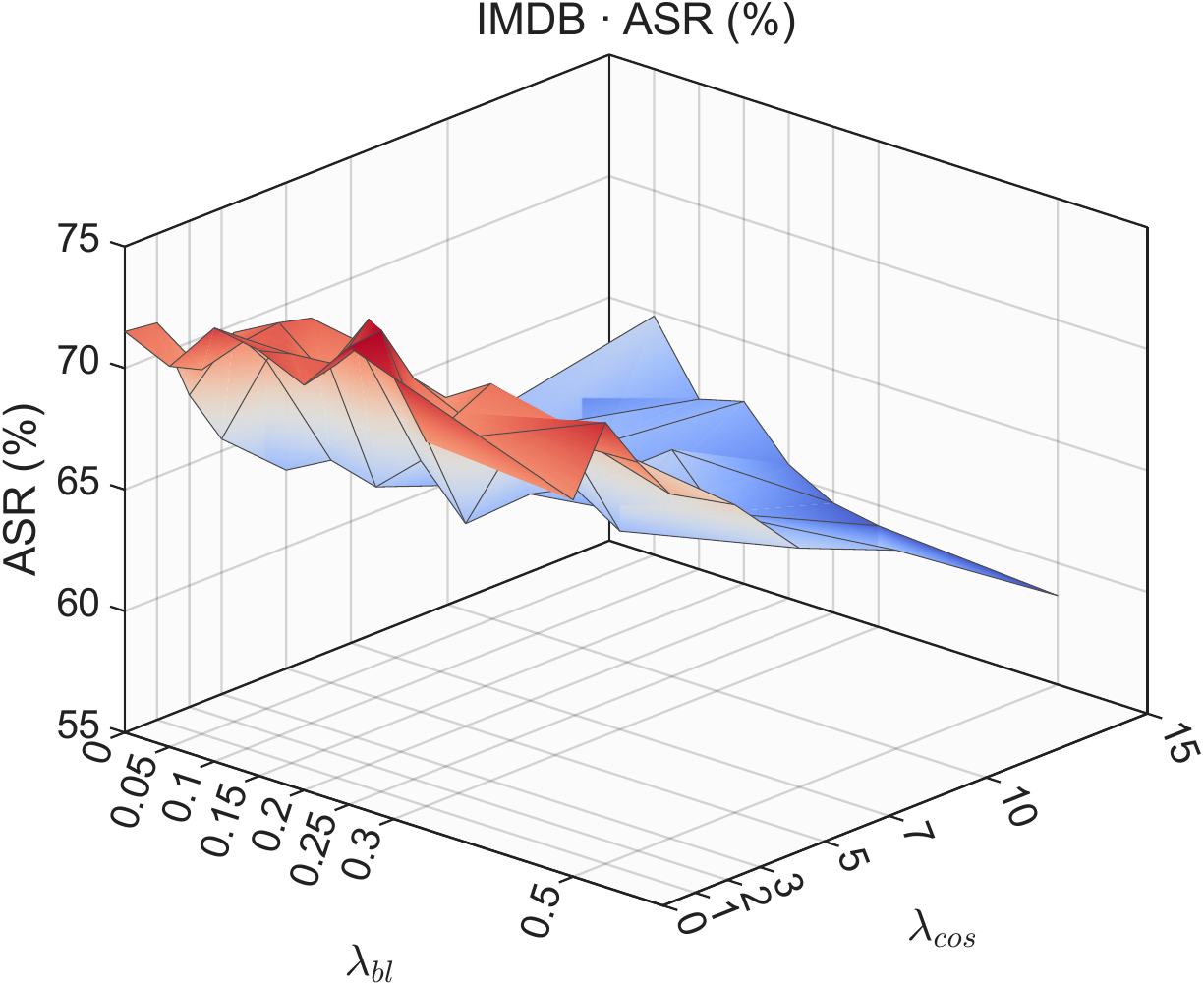}
   \hfill
   \includegraphics[width=0.24\textwidth]{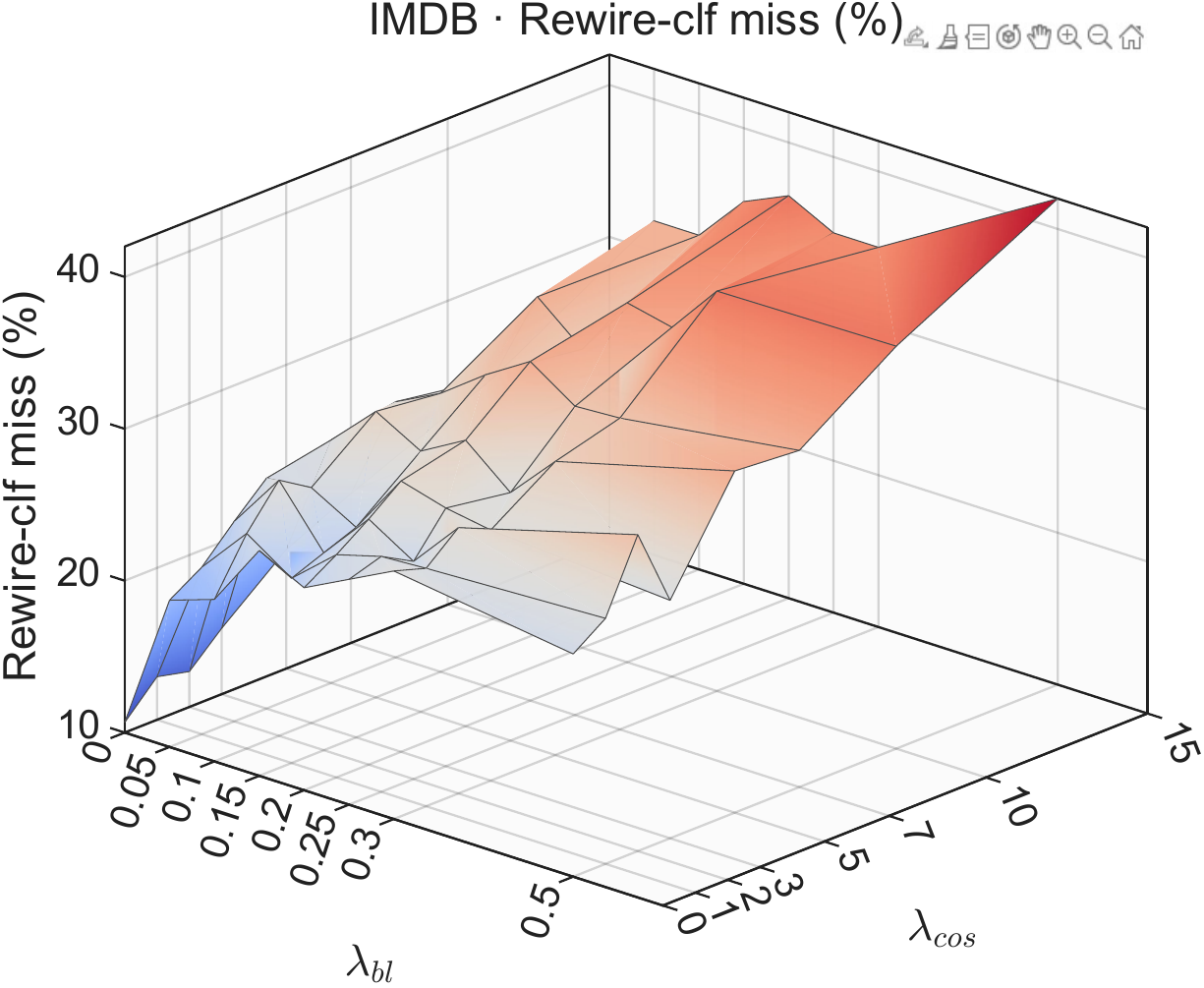}
   \caption{Hyperparameter analysis of Blackknife on DBLP and IMDB. From left to right: DBLP attack success rate, DBLP classifier miss rate, IMDB attack success rate, and IMDB classifier miss rate.}
   \label{fig:hyperparameter_dblp_imdb}
\end{figure*}

We analyze the effects of the structural classifier weight $\lambda_{\mathrm{bl}}$ and cosine similarity weight $\lambda_{\mathrm{cos}}$, varying them from $0$ to $0.5$ and from $0$ to $15$, respectively. Fig.~\ref{fig:hyperparameter_dblp_imdb} reports ASR and classifier miss rate on DBLP and IMDB. Classifier miss rate is the fraction of attacked examples classified as clean by the independent evaluation detector; a higher value indicates better evasion.

On DBLP, weaker regularization generally achieves higher ASR, while some combinations with stronger regularization increase the classifier miss rate at the cost of attack effectiveness. These trends are not strictly monotonic, and the settings with high miss rates do not necessarily yield high ASR. On IMDB, ASR declines more clearly as $\lambda_{\mathrm{cos}}$ increases, while classifier evasion improves in several regions. These surfaces illustrate the trade-off between attack effectiveness and detection evasion. Additional results on ACM are provided in Fig.~\ref{fig:hyperparameter_acm_3d} in Appendix~\ref{app:hyperparam}.

\subsection{Degree Distribution Shift (RQ5)}
\begin{figure*}[t]
   \centering
   \includegraphics[width=0.32\textwidth]{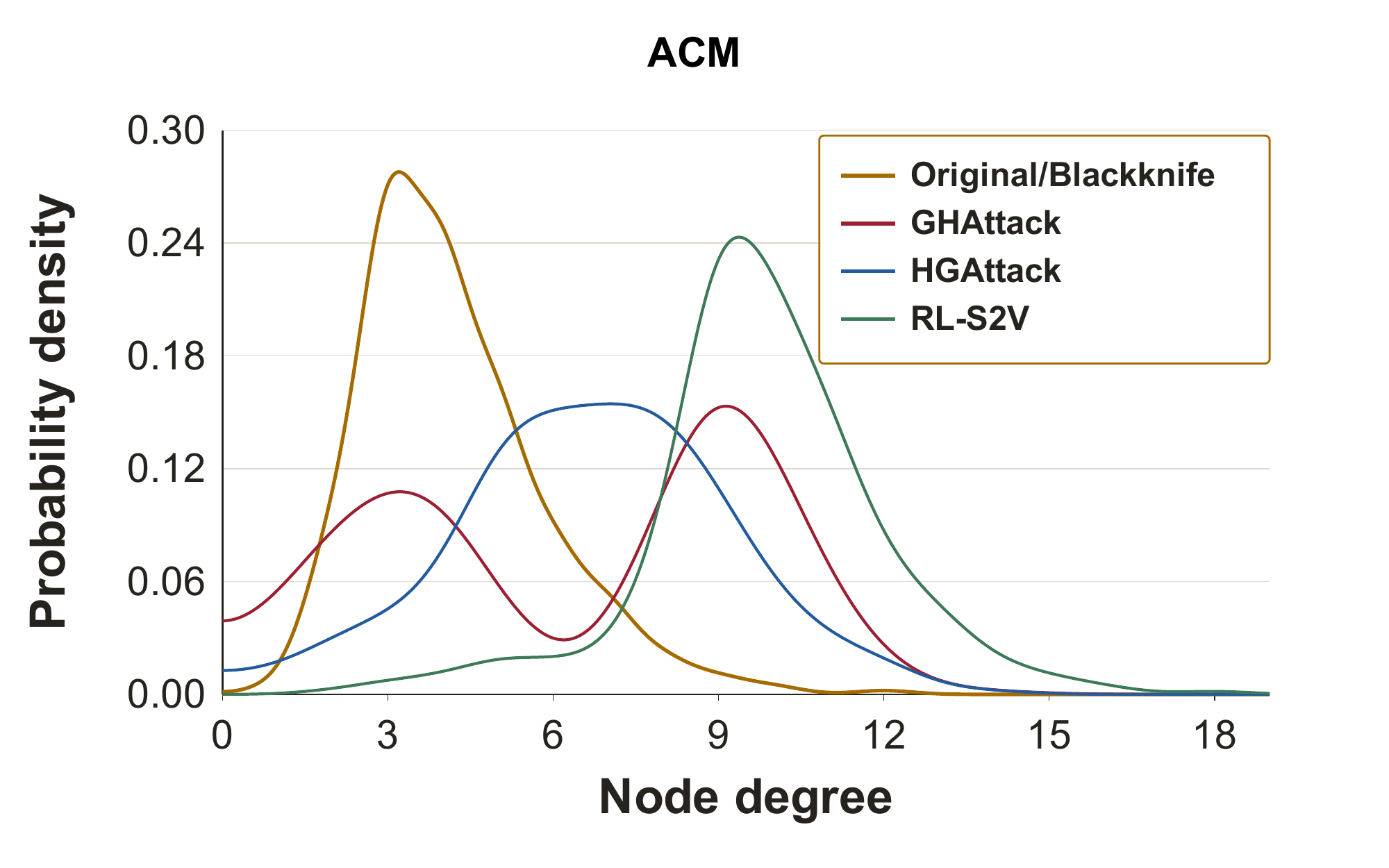}
   \hfill
   \includegraphics[width=0.32\textwidth]{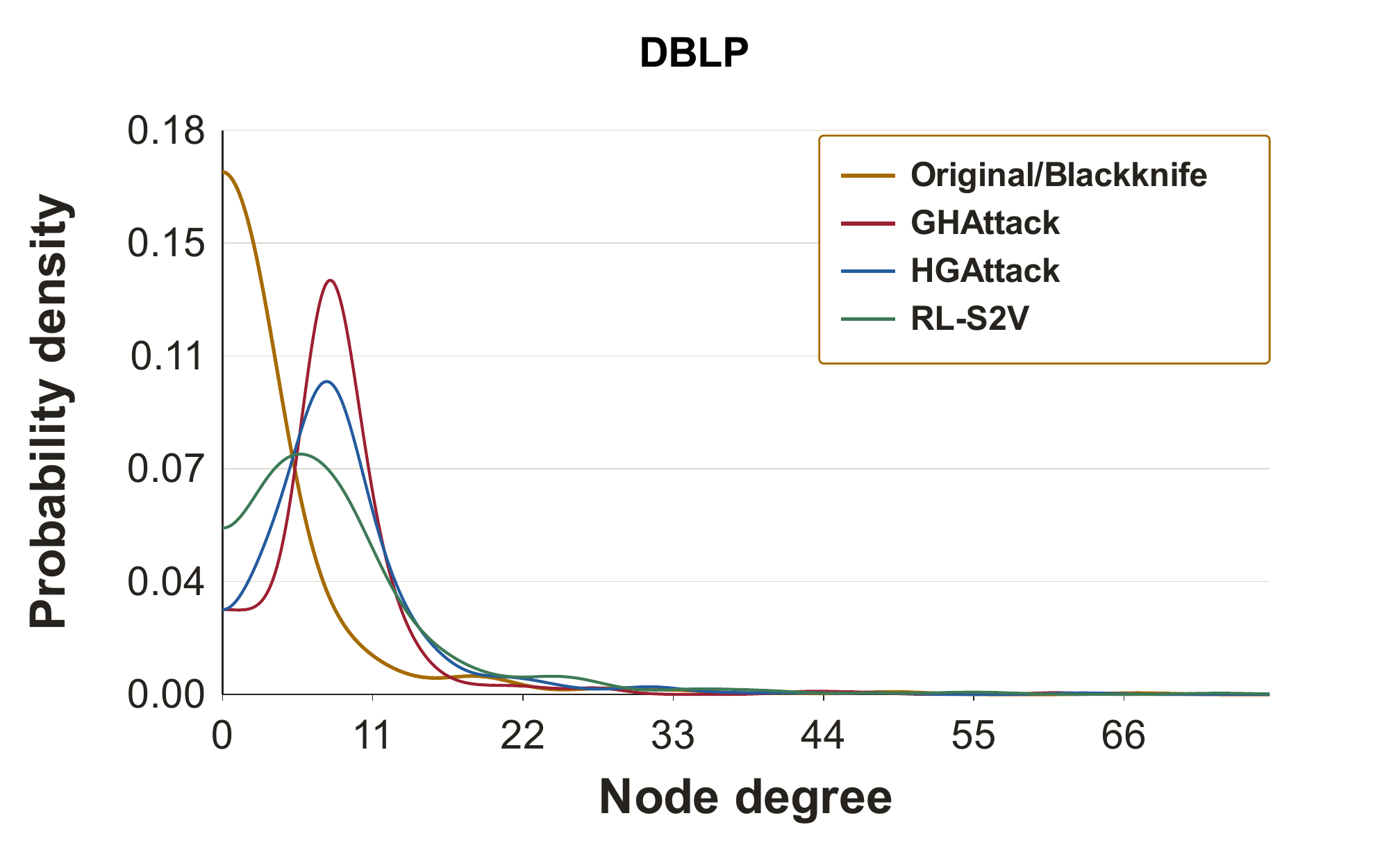}
   \hfill
   \includegraphics[width=0.32\textwidth]{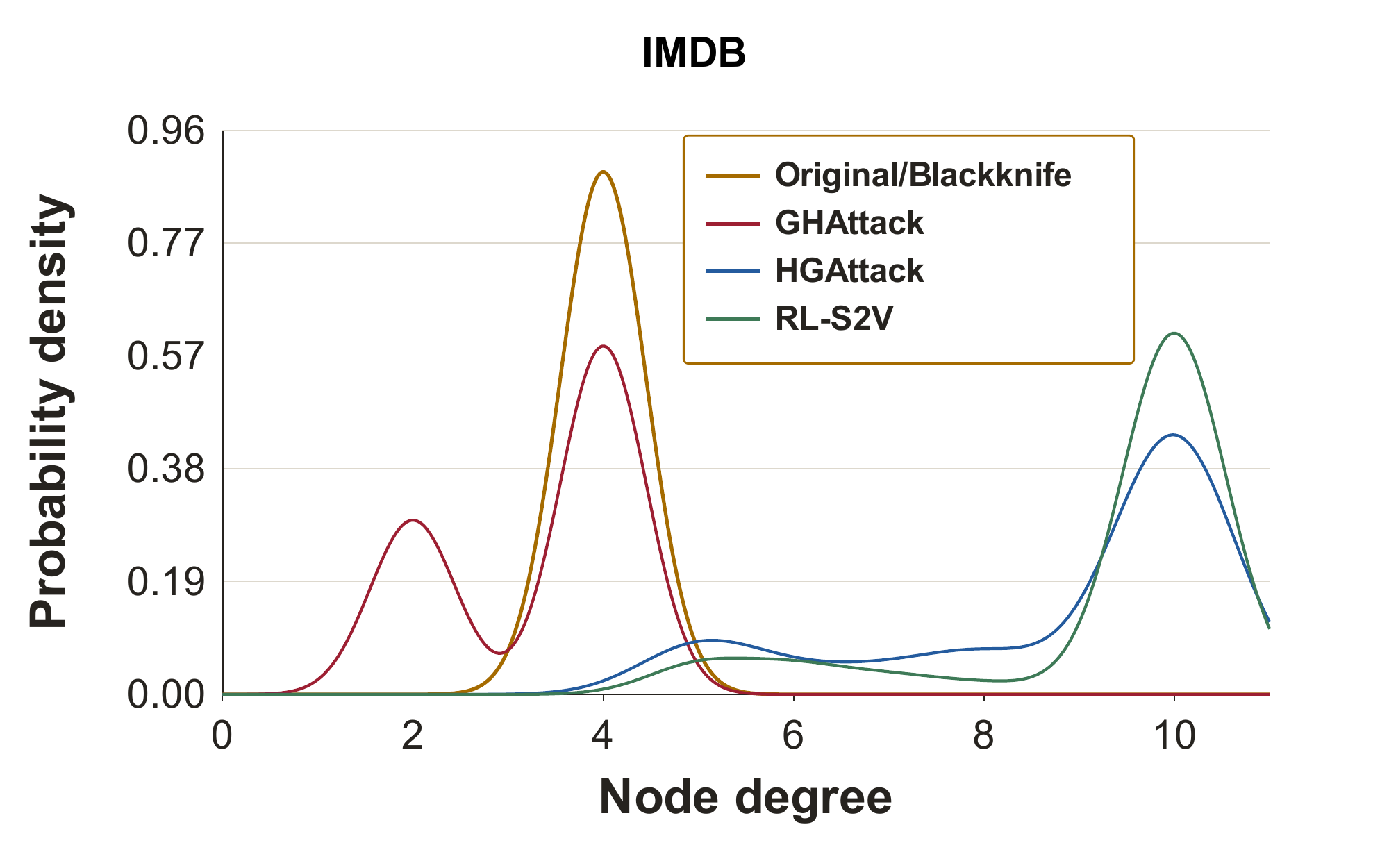}
   \caption{Degree distributions of attacked target nodes before and after separate attacks on ACM (left), DBLP (middle), and IMDB (right).}
   \label{fig:degree_distribution_shift}
\end{figure*}

Fig.~\ref{fig:degree_distribution_shift} compares target node degrees before and after the baseline attacks. Independent edge additions and deletions can shift the target degree distribution, as shown by the baseline curves. These shifts distort the original degree profile and introduce additional structural signals for attack detection.

Blackknife pairs each deletion with an addition of the same relation type, preserving every target’s total degree and degree for each relation type. Its target degree distribution therefore remains identical to the clean distribution. This preserves the original degree profile while changing neighborhood composition, avoiding the structural distortion caused by changes in target degrees.

\section{Limitations and Future Work}
\label{sec:limitations}
Blackknife has two main limitations. First, its local surrogate cannot capture global structural changes beyond the observed one hop neighborhoods. This limits its ability to account for defenses that purify graph structures using information outside its observation scope, such as TGNN \cite{TGNN} and DiffSP \cite{DiffSP}.

Second, although Blackknife requires only one hop structural observations, it still relies on observations and labels from a batch of target nodes to train its shared surrogate. This dependence limits its applicability when only a few target nodes are observable. Future work will investigate how to account for global structural changes under restricted observations and train effective surrogates from fewer observed targets.

\section{Conclusion}
\label{sec:conclusion}
We propose Blackknife, a black box evasion attack on HGNNs that generates and selects perturbations without querying the victim model. A shared local surrogate uses one hop structural observations, known target labels, and permitted raw features to guide rewiring while preserving each target’s total degree and degree for each relation type. Cosine similarity regularization and an adaptive structural classifier support detection evasion.

Experiments demonstrate Blackknife’s effectiveness across different victim architectures and defense settings. These findings show that changing neighborhood composition can compromise HGNN predictions under restricted structural access, even when target degrees are preserved and attack generation proceeds without victim feedback.

\printcredits

\bibliographystyle{cas-model2-names}

\bibliography{cas-refs}

\appendix
\section{Other Hyperparameter Analysis}
\label{app:hyperparam}
\begin{figure}[pos=!htbp]
   \centering
   \includegraphics[width=0.48\columnwidth]{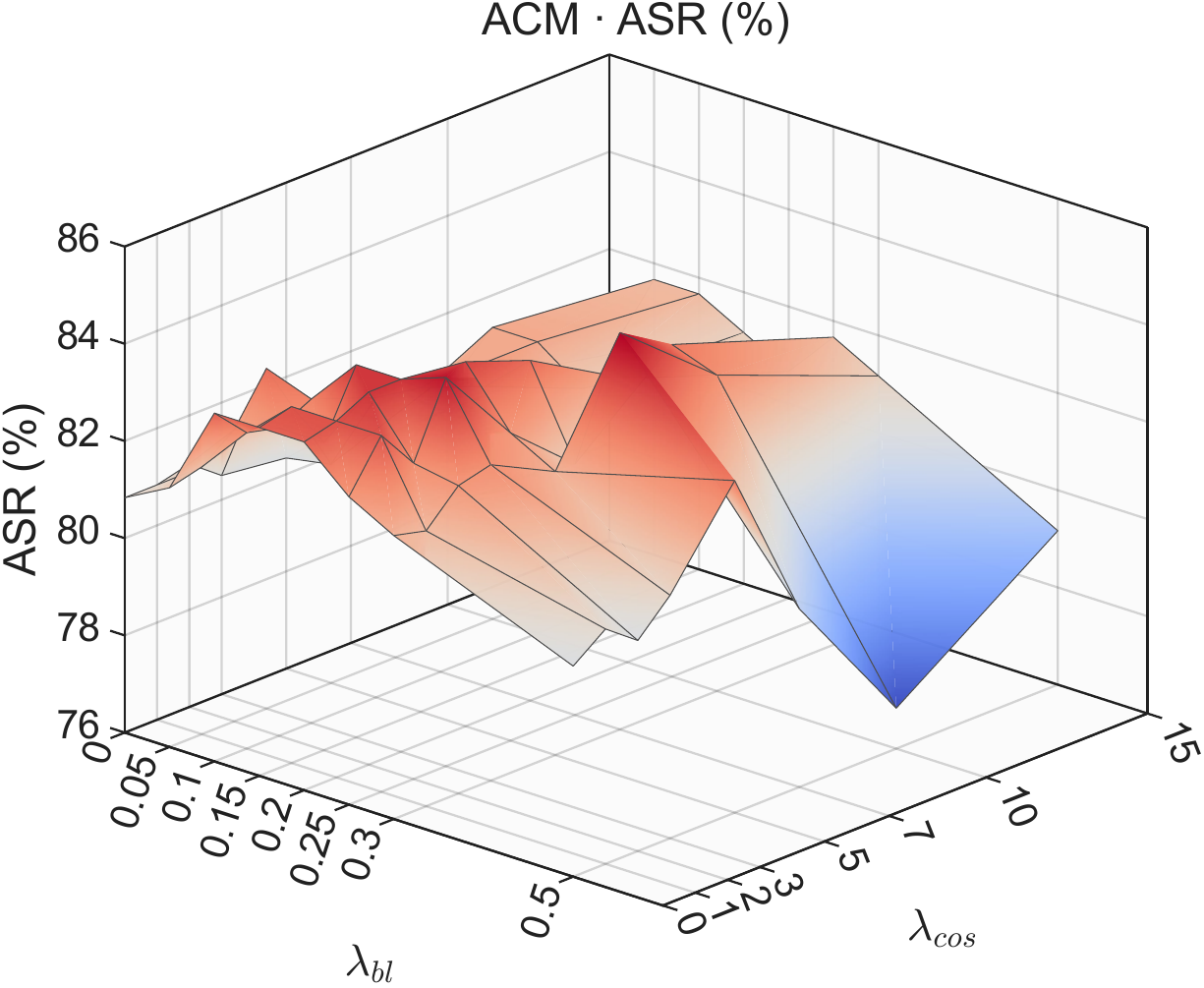}
   \hfill
   \includegraphics[width=0.48\columnwidth]{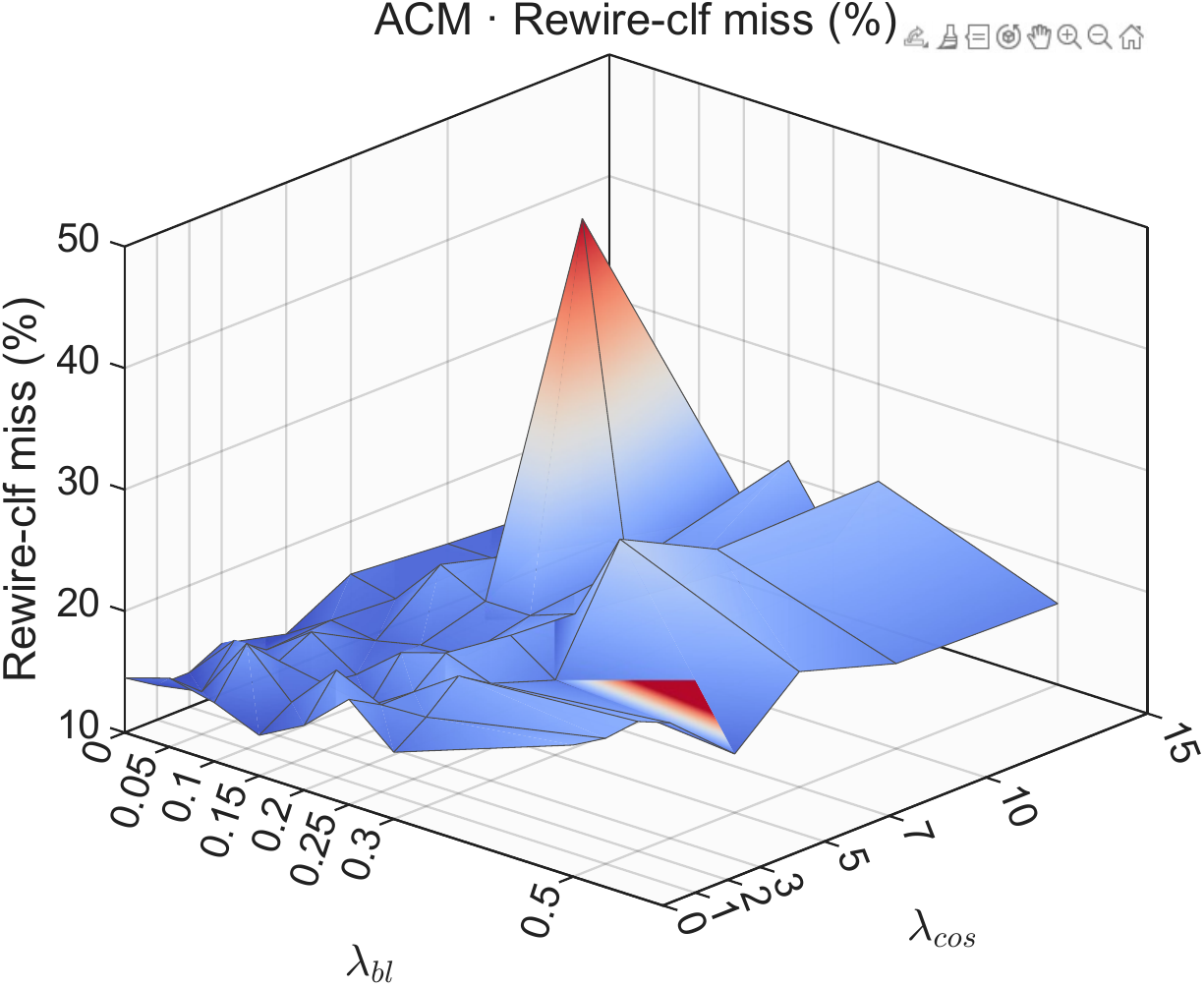}
   \caption{Hyperparameter analysis of Blackknife on ACM: attack success rate (left) and classifier miss rate (right).}
   \label{fig:hyperparameter_acm_3d}
\end{figure}

Fig.~\ref{fig:hyperparameter_acm_3d} reports the hyperparameter analysis on ACM. ACM maintains relatively high ASR across many settings, with a localized peak in classifier miss rate.

\section{Proofs of the Theoretical Results}
\label{app:proofs}
\begin{proof}[Lemma 1]
By definition, the normalized propagation matrix $P_H$ is constructed entirely from the relation-specific adjacency matrices within the fixed local subgraph $G_{v_p}$. Meanwhile, the initial representation matrix $\widetilde{Z}$ contains only the representations of nodes in $G_{v_p}$. Since the representation after $L$ propagation layers is given by $H^{(L)}=P_H^L\widetilde{Z}$, both $P_H^L$ and $\widetilde{Z}$ depend exclusively on $G_{v_p}$. Therefore, for any propagation depth $L$, $H^{(L)}$ depends only on the fixed local subgraph $G_{v_p}$ and does not require any information outside the observable scope.
\end{proof}

\begin{proof}[Theorem 1]
The matrix $D_H$ records the row sums of $B_H$. Normalizing $B_H$ by $D_H^{-1}$ therefore ensures that each row of $P_H=D_H^{-1}B_H$ sums to one. Moreover, since all entries of $B_H$ are nonnegative, $P_H$ is a row-stochastic matrix and can therefore be interpreted as the transition matrix of a finite-state Markov propagation process over the nodes in $G_{v_p}$.

Under the assumption that this propagation process is irreducible and aperiodic, the standard convergence theorem for finite Markov chains guarantees the existence of a unique stationary distribution $\pi_H$ \cite{markov}. Moreover, as $L\to\infty$, every row of $P_H^L$ converges to this stationary distribution. In particular, for the target node $v_p$, we have $\lim_{L\to\infty}[P_H^L]_{v_p,:}=\pi_H^{\mathsf{T}}$.

Since $H^{(L)}=P_H^L\widetilde{Z}$ and $h_{v_p}^{(L)}=[H^{(L)}]_{v_p,:}$, the representation of the target node satisfies $h_{v_p}^{(L)}=[P_H^L]_{v_p,:}\widetilde{Z}$. Taking the limit as $L\to\infty$ gives $\lim_{L\to\infty}h_{v_p}^{(L)}=\pi_H^{\mathsf{T}}\widetilde{Z}=\sum_{u\in G_{v_p}}\pi_H(u)\widetilde{z}_u$. Therefore, repeated propagation within the fixed local subgraph converges to a stable limiting representation.
\end{proof}

\begin{proof}[Theorem 2]
By $H^{(L)}=P_H^L\widetilde{Z}$, the target representation is
\[
h_{v_p}^{(L)}
=\sum_{u\in\mathcal{U}}[P_H^L]_{v_p,u}\widetilde{z}_u.
\]
Expanding the matrix product gives
\[
[P_H^L]_{v_p,u}
=\sum_{x_1,\ldots,x_{L-1}\in\mathcal{U}}
P_H(v_p,x_1)\cdots P_H(x_{L-1},u).
\]
Each nonzero term corresponds to a length-$L$ directed walk within
the local support graph, with weight equal to the product of its
transition weights. Hence $[P_H^L]_{v_p,u}=\alpha_{v_p,u}^{(L)}$.
For $L=1$, this identity reduces to
$\alpha_{v_p,u}^{(1)}=P_H(v_p,u)$.
All intermediate nodes belong to $\mathcal{U}$ and may repeat, so
the expansion holds for every finite $L$, including depths exceeding
the observation radius. The same argument applies to $P'_H$.

The difference between the target representations is $h_{v_p}^{(L)}-h_{v_p}^{\prime(L)}=\bigl([P_H^L]_{v_p,:}-[(P'_H)^L]_{v_p,:}\bigr)\widetilde{Z}$.
The coefficient difference is nonzero by assumption. Because the
rows of $\widetilde{Z}$ are linearly independent, a nonzero linear
combination of these rows is nonzero. Consequently,
$h_{v_p}^{(L)}\neq h_{v_p}'^{(L)}$.
\end{proof}

\section{Node Classification with Local Subgraphs}
\label{app:surrogate}
\begin{table}[ht]
   \centering
   \small
   \caption{Classification accuracy of an independent local-context model on the standard dataset splits.}
   \label{tab:surrogate_performance}
   \resizebox{\linewidth}{!}{%
\begin{tabular}{lccc}
   \toprule
   \textbf{Dataset} & \textbf{Train} & \textbf{Val} & \textbf{Test} \\
   \midrule
   ACM  & $0.9951\pm0.0044$ & $0.9156\pm0.0043$ & $0.9175\pm0.0076$ \\
   IMDB & $0.9968\pm0.0000$ & $0.5917\pm0.0106$ & $0.5641\pm0.0023$ \\
   DBLP & $0.9914\pm0.0012$ & $0.7841\pm0.0051$ & $0.7947\pm0.0038$ \\
   \bottomrule
   \end{tabular}
}
\end{table}
Table~\ref{tab:surrogate_performance} reports an auxiliary classification experiment using one-hop local contexts. Test accuracy is $0.9175$ on ACM, $0.7947$ on DBLP, and $0.5641$ on IMDB. These results show that local contexts contain useful class information and support node classification across different datasets.

\end{document}